\documentclass[a4paper,fleqn]{cas-dc}

\usepackage[authoryear,longnamesfirst]{natbib}
\usepackage{caption}
\usepackage{subfigure}
\usepackage{amssymb}
\usepackage{bm}
\usepackage[ruled, linesnumbered]{algorithm2e}
\usepackage{amsmath}
\usepackage{mathtools}
\DeclarePairedDelimiter{\ceil}{\lceil}{\rceil}

\def\tsc#1{\csdef{#1}{\textsc{\lowercase{#1}}\xspace}}
\tsc{WGM}
\tsc{QE}

\begin{document}
\let\WriteBookmarks\relax
\def\floatpagepagefraction{1}
\def\textpagefraction{.001}



\title [mode = title]{SSD-KD: A Self-supervised Diverse Knowledge Distillation Method for Lightweight Skin Lesion Classification Using Dermoscopic Images}  




  \author[1]{Yongwei Wang}
   \fnmark[1] 
  \credit{Conceptualization, Methodology, Data Curation, Software, Writing - Original Draft, Writing - Review \& Editing}
  \author[2,3,4,5,6]{Yuheng Wang}
  [orcid = 0000-0001-5252-7005]
  \cormark[1] 
  \fnmark[2] 
  \ead{yuhengw@ece.ubc.ca}
    \credit{Conceptualization, Methodology, Formal analysis, Validation, Writing - Original Draft, Writing - Review \& Editing}    
  \author[3,4,5,6] {Tim K. Lee}
  [orcid = 0000-0002-0157-2081]
    \credit{Supervision, Project administration, Funding acquisition, Writing - Review \& Editing}
  \author[7]{Chunyan Miao}
  \cormark[1]
  \ead{ascymiao@ntu.edu.sg}
     \credit{Supervision, Project administration, Funding acquisition, Writing - Review \& Editing}
  \author[2,3]{Z. Jane Wang}
  [orcid = 0000-0002-3791-0249]
    \credit{Supervision, Project administration, Funding acquisition, Writing - Review \& Editing}
  
  \address[1]{Joint NTU-WeBank Research Center on Fintech, Nanyang Technological University, Singapore}   
  \address[2]{Department of Electrical and Computer Engineering, University of British Columbia, Vancouver, BC, Canada}
  \address[3]{School of Biomedical Engineering, University of British Columbia, Vancouver, BC, Canada}
  \address[4]{Department of Dermatology and Skin Science, University of British Columbia, Vancouver, BC, Canada }
  \address[5]{Photomedicine Institute, Vancouver Coast Health Research Institute, Vancouver, BC, Canada }
  \address[6]{Cancer Control Research Program, BC Cancer, Vancouver, BC, Canada}
  \address[7]{School of Computer Science and Engineering, Nanyang Technological University, Singapore}

  \cortext[cor1]{Corresponding author.} 
  \fntext[fn1]{Yongwei Wang did part of the work at UBC, and he has moved to NTU.}
  \fntext[fn2]{Yongwei Wang and Yuheng Wang contributed equally to this paper.}




\begin{abstract}
Skin cancer is one of the most common types of malignancy, affecting a large population and causing a heavy economic burden worldwide. Over the last few years, computer-aided diagnosis has been rapidly developed and make great progress in healthcare and medical practices due to the advances in artificial intelligence, particularly with the adoption of convolutional neural networks. However, most studies in skin cancer detection keep pursuing high prediction accuracies without considering the limitation of computing resources on portable devices. In this case, the knowledge distillation (KD) method has been proven as an efficient tool to help improve the adaptability of lightweight models under limited resources, meanwhile keeping a high-level representation capability. To bridge the gap, this study specifically proposes a novel method, termed SSD-KD, that unifies diverse knowledge into a generic KD framework for skin diseases classification. Our method models an intra-instance relational feature representation and integrates it with existing KD research. A dual relational knowledge distillation architecture is self-supervisedly trained while the weighted softened outputs are also exploited to enable the student model to capture richer knowledge from the teacher model. To demonstrate the effectiveness of our method, we conduct experiments on ISIC 2019, a large-scale open-accessed benchmark of skin diseases dermoscopic images. Experiments show that our distilled lightweight model can achieve an accuracy as high as 85\% for the classification tasks of 8 different skin diseases with minimal parameters and computing requirements. Ablation studies confirm the effectiveness of our intra- and inter-instance relational knowledge integration strategy. Compared with state-of-the-art knowledge distillation techniques, the proposed method demonstrates improved performances. To the best of our knowledge, this is the first deep knowledge distillation application for multi-diseases classification on the large-scale dermoscopy database. Our codes and models are available at: \url{https://github.com/enkiwang/Portable-Skin-Lesion-Diagnosis}.
\end{abstract}




\maketitle

\section{Introduction}\label{intro}

Skin cancer is among the most common human malignancies globally, especially among the fair-skinned population. Over the past decades, skin cancer has pervaded different cultures and caused a huge economic burden in the healthcare maintenance. Meanwhile, both malignant melanoma (MM), which causes most deaths in skin cancers, and keratinocyte carcinomas (squamous cell carcinoma and basal cell carcinoma) have a promising prognosis if they are detected and well treated at the early stage \citep{apalla2017epidemiological}. Therefore, early diagnosis plays an essential role in skin cancer’s efficient management and successful treatment.

Benefiting from the unprecedented advances in optical imaging techniques, huge amounts of high-quality skin images were collected in multiple modalities. Dermoscopy, a noninvasive digital imaging technique, has been widely used for skin cancer diagnosis since it allows the in vivo evaluation of colors and structures of the epidermis that are hard to be visualized by the naked eye \citep{celebi2019dermoscopy,barata2018survey}. In traditional skin cancer diagnosis, expert dermatologists can inspect the skin lesions following valid algorithms such as ABCD rules \citep{nachbar1994abcd}, 7-point Checklist \citep{argenziano1998epiluminescence} using dermoscopy images. However, such time consuming and expensive diagnostic methods cause a great burden on clinical diagnosis. Therefore, automatic skin cancer detection is in urgent need to relieve dermatologists’ heavy work pressure in clinical diagnosis.

Convolutional neural networks (CNN) have been widely used in computer-aided diagnosis (CAD) for skin cancer detection and achieved remarkable performance, because of its excellent and robust capability in feature extraction and categorical classification \citep{esteva2017dermatologist,litjens2017survey,brinker2018skin}. In recent years, more and more research efforts have been made to develop advanced deep CNN architectures to improve their diagnostic performance for skin cancer detection. Esteva et al. \citep{esteva2017dermatologist} involved the pre-trained Inception V3 models in skin lesions’ classification and presented dermatologist-leveled prediction results for the first time. Abbas et al. \citep{abbas2019dermodeep} proposed the DermoDeep framework for the classification of melanoma and nevus that consisted visual-features fusion and deep learning without pre- and post-processing steps.  Kawahara et al. \citep{kawahara2016deep} found that the features extracted from the ImageNet pre-trained model outperformed general handcrafted features in skin lesion classification tasks. With the further development of deep learning technology, more novel structures have also been proposed and being used in this task. The integration strategies of various deep learning models and the interpretive analysis of the predictions they produce have also been extensively studied. For example, Gessert et al. \citep{gessert2020skin} proposed an ensemble framework for skin lesion classification using multi-resolution Efficient Net architectures with a search strategy. Barata et al. \citep{barata2021explainable} came up with an explainable framework that can provide insightful information about deep learning models’ decisions by incorporating the taxonomies information and special attention blocks. Meanwhile, the exploration of the combined effect of multiple modalities and the introduction of clinical diagnostic knowledge have further promoted the development of deep learning in the diagnosis of skin diseases. Kawahara et al. \citep{kawahara2018seven} proposed a multi-task deep convolutional neural network containing clinical, dermoscopic images and patient metadata. Their approach incorporated clinical rule-based knowledge into deep learning and demonstrated benchmark results for 7-point criteria and diagnostics. Based on this study, Bi et al. \citep{bi2020multi} further considered the feature information that multi-modal input can share in the training phase of deep learning architecture and achieved further improvement in melanoma detection. Pacheco et al. \citep{pacheco2021attention} proposed a MetaBlock module to improve classifiers' performance by utilizing the clinical metadata. Wang et al. \citep{wang2021incorporating} proposed a novel method that introduced the inter-dependencies between different criteria using a constrained classifier chain, which made the resulting predictions more acceptable to physicians. Tang et al. \citep{tang2022fusionm4net} came up with a new algorithm that fused both feature-level and decision-level information from multiple modalities and achieved state-of-the-art performance in skin cancer diagnosis. 

These remarkable achievements using varieties of CNN architectures have a great impact on the progress of automated skin cancer diagnosis. However, these research works related to CNNs also indicate that sophisticated models with more neural network layers and blocks tend to have better predictive performance with sufficient training data and proper training procedures. These works further suggest that, CNNs necessitate a huge number of parameters and massive floating point operations to obtain a satisfactory performance. For example, ResNet50 \citep{he2016deep}, as one of the most well-known and effective deep learning  models, has about 25.6 millions of parameters, which requires 98 MB and 4.11 GFlops to process the algorithm. Although these complex and powerful deep learning models can allow us to obtain better prediction results, they also require significant computing resources (e.g., memories and hardware resources). These strict requirements may impede their practicability in restrained environments and the embedded devices that have limited computation power and memories. Therefore, exploring portable and efficient networks with comparable performance for skin cancer auto-detection that can work on embedded equipment or limited resources has both practical and theoretical significance.

Knowledge Distillation (KD), proposed as an extended model compression method, can transfer the learned knowledge from a complicated model (the teacher) to a simple model (the student) by sharing the soft labels during the training phase, thus improving the representation power of the lightweight CNNs preferably \citep{hinton2015distilling, tung2019similarity}. In recent years, many novel and practical deep learning models based on KD ideas have been proposed. Romero et al. \citep{romero2015fitnets} proposed the FitNet, an approach that added an additional requirement of predicting the outputs of the intermediate layers to the traditional KD algorithm and made it suitable for thin and deep networks. Yim et al. \citep{yim2017gift} measured a flow of feature correlations among adjacent layers in residual blocks and transferred them to a student network, then Liu et al. \citep{liu2019knowledge} extended it to a dense cross-layer transfer scenario. Park et al. \citep{park2019relational} built a dubbed relational KD deep learning framework that considered both distance-wise and angle-wise distillation losses and improved the prediction ability of student models significantly. Peng et al. \citep{peng2019correlation} proposed a correlation congruence distillation scheme to model the relation between samples in the kernel form. Xu et al. \citep{xu2020knowledge} further found that self-supervision signals can effectively transfer the hidden information from the teacher to the student via their proposed SSKD architecture, which substantially benefited the scenarios with few-shot and noisy-labels. Ding et al. \citep{ding2021distilling} proposed an data augmentation-based  knowledge distillation framework for classification and regression tasks via synthetic samples \citep{ding2020ccgan,ding2021efficient}. 

Meanwhile, KD has been proven effective in the compression of deep learning network structures on the basis of ensuring model prediction and learning capabilities among varieties of areas \citep{gou2021knowledge,wang2021knowledge}. In particular, KD shows great potential for applications in computer-aided diagnostics. For example, Qin et al. \citep{qin2021efficient} proposed an efficient architecture by distilling knowledge from well-trained medical image segmentation networks to train a lightweight network and obtained significant improvements for the segmentation capability while retaining the runtime efficiency. Abbasi et al. \citep{abbasi2021classification} implemented knowledge distillation on unlabeled medical data via an unsupervised learning manner and showed a significant improvement in the prediction of diabetic retinopathy. Chen et al. \citep{chen2022lightweight} compared multiple deep learning models’ performance regarding cervical cells classification using knowledge distillation and showed the importance of model selection.  At the same time, knowledge distillation methods have also shown great potential in the field of skin disease diagnosis although there still exists quite limited attention in this area. Back et al. \citep{back2021robust} used an ensemble learning strategy to help a student network learn from multiple teacher networks progressively in the portable diagnosis of herpes zoster skin disease using clinical images, and achieved robust performances. To our best knowledge, however, there are still no studies on the classification of multiple skin diseases based on large-scale dermocopy image datasets.

\begin{figure*}
\centering
\includegraphics[width=1\textwidth]{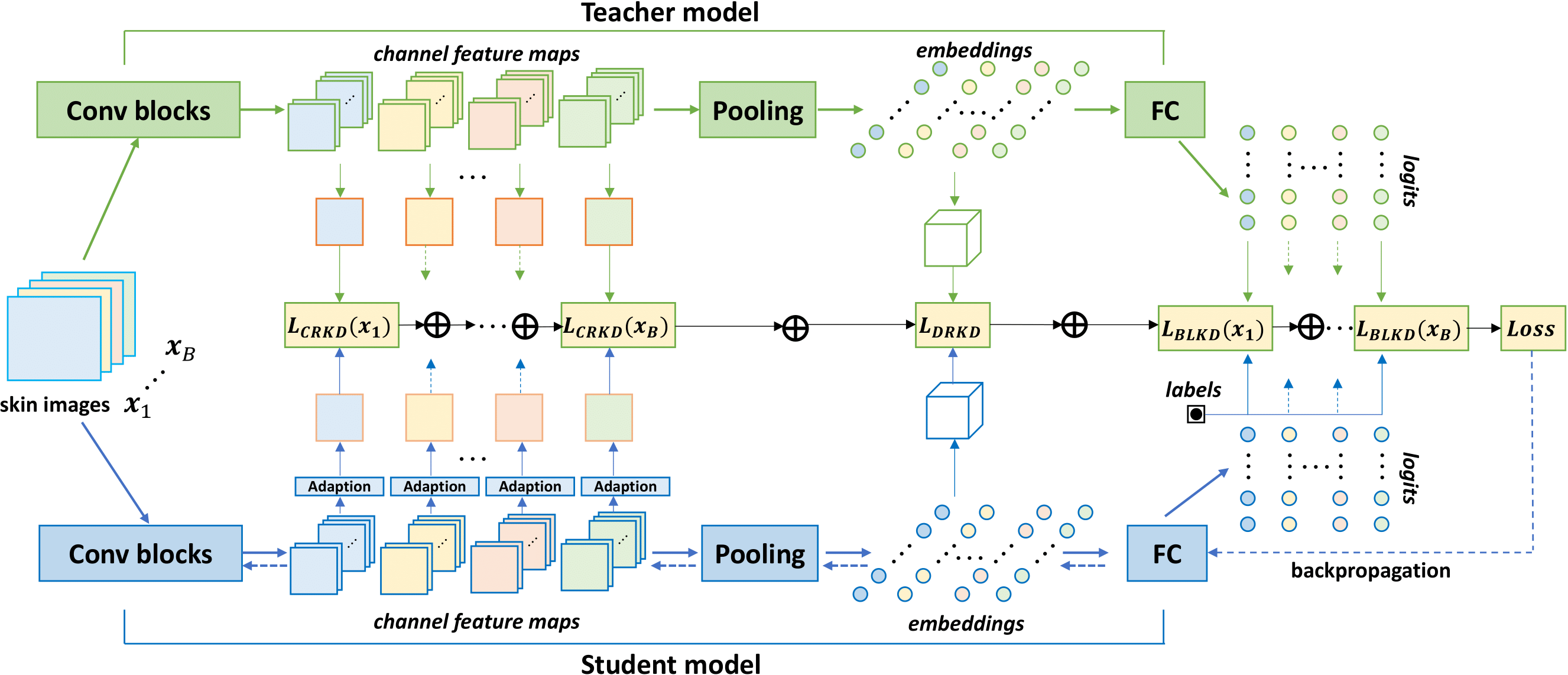}
\caption{\label{fig:flowchart}An illustration of the proposed knowledge distillation framework that distills and transfers diverse knowledge to the student model for skin cancer diagnosis. The upper branch (in green) denotes the pretrained teacher classification model, and the lower branch (in blue) denotes the untrained student classification model. Both models consist of a set of convolutional blocks to extract feature maps of a batch of skin images, followed by average pooling operations to produce embeddings, and a fully connected (FC) layer to yield logits for skin disease classification. The three types of knowledge and corresponding loss functions are quantified based on channel feature maps, embeddings and logits, respectively. In particular, our novel channel relational distillation loss is defined in the feature map stream. Finally, the diverse knowledge-integrated distillation loss is computed and backpropagated (along dashed reverse arrow lines) to train the student network. The self-supervised learning strategy can be conveniently incorporated to further boost the training of the student model. }
\end{figure*}

Despite that these innovative studies have advanced the application of KD in the field of computer-aided diagnosis, they still mainly rely on the sharing and transferring the subject-level category information between the teacher network and the student network. In detail, existing skin diagnosis KDs have largely ignored the inter and intra relations within input samples, which may preserve important knowledge to diagnose skin diseases from the teacher network. This work explores the intra-instance relational knowledge, as well as its integrated analysis with traditional inter-instance relational knowledge and dark knowledge, to enhance the effectiveness of information transfer from the teacher network to the student network. Consequently, the proposed diverse knowledge-aware KD can effectively improve the predictive capabilities of the student model.

In this work, we develop a novel knowledge distillation framework, termed SSD-KD, that integrates diverse conventional knowledge with a novel type of intra-instance relational knowledge to improve the auto-diagnosis of multiple types of skin diseases based on lightweight deep learning models. Specifically, we formulate a dual relational knowledge distillation architecture introducing different relational representations and softened network outputs to distill and transfer diverse knowledge from the teacher model to student model. Moreover, we also incorporate the self-supervised knowledge distillation strategy into the framework to enable the student model capture richer structured knowledge from the self-supervision predictions of the teacher model. Additionally, to make the proposed SSD-KD framework more suitable for the skin lesions auto-detection task, which always suffers from data imbalance problems, we replace the general cross-entropy loss function in conventional KDs with an improved weighted version to benefit the skin disease categories with less subjects. We have evaluated our proposed method in the publicly available skin lesion datasets ISIC 2019, which contains over 20000 dermoscopy images and 8 types of skin diseases. The comparison with state-of-the-art methods and ablation studies are also performed to demonstrate the effectiveness of our proposed framework. 

The main contributions of this study can be summarized as follows: \par

\begin{itemize}
     \item We propose a novel self-supervised diverse knowledge distillation method for lightweight multi-class skin diseases classification using dermoscopy images, named SSD-KD. We replace the traditional single relational modeling block with dual relational blocks in our method, which assists the student model capture both intra- and inter-instance relational information from the teacher model. Hence, our diverse knowledge-integrated KD can further improve the transferable ability of KD and the performance for skin disease classification.
     \item We employ the self-supervision based auxiliary learning strategy in our knowledge distillation framework, which enables richer structured knowledge to be transferred, and stabilizes the training dynamics of the intra- and inter-instance relational distillation scheme.
     \item We utilize the weighted cross entropy loss function and combine it with our SSD-KD structure to assist the lightweight student model better mimic the behavior of the teacher model in capturing additional information from the diseases with limited subjects. 
     \item The results demonstrate that our carefully designed method can help the student model achieve comparable or even better performance than the teacher model in the multi-class skin lesion classification task. From a practical clinical point of view, our proposed method can further advance the application of remarkable deep learning models in scenarios with limited computing resources.
\end{itemize}\par
The rest of the paper is organized as follows: the detailed introduction of the proposed method is presented in the second section, followed by the experiment setup and results analysis in the third section. The discussion and the future works are shown in the fourth section while the conclusion of the study is put in the last section.

\section{Methodology}

In this section, we describe the proposed method that aims to develop an effective knowledge distillation framework for skin disease detection. The overall pipeline of the proposed method is shown in Figure~\ref{fig:flowchart}. In the proposed framework, we distill and transfer three types of knowledge from a large and pretrained teacher model to a portable student model that will be trained from scratch. The first type of knowledge adopts a customized version of the conventional knowledge distillation scheme \citep{hinton2015distilling} that defines the ``dark knowledg'' via logits (i.e., the output vector from the last layer of a CNN model). The second type of knowledge utilizes the inter-instance relational knowledge \citep{park2019relational} to model the relationship of a batch of training samples at the penultimate layer.  In addition, we define the intra-instance relational knowledge between channel feature maps of each sample at the last convolutional layer.  This type of knowledge further exploits the inherent structures within model responses to facilitate the training of the student model.

For convenience, we introduce some notations to be used in the following section. Let us denote the overall training set of skin images as $\mathcal{X}$ with groundtruth labels as $\mathcal{Y}$. Following the general routine in deep learning, we split the training set $(\mathcal{X}, \mathcal{Y})$ into non-overlapping subsets randomly. The number of skin images (i.e., batch size) in each subset is denoted as $B$, i.e., $\{ (\boldsymbol{x}_1, y_1), \cdots,  (\boldsymbol{x}_B, y_B) \} \subset \mathcal{X}$. For an input sample $\boldsymbol{x}_i \ (i=1,\cdots, B)$, we denote its logits vector by $\boldsymbol{l}(\boldsymbol{x}_i)$, the embedding vector by $\boldsymbol{e}_i$ (i.e., $\boldsymbol{e} (\boldsymbol{x}_i)$ for brevity), and channel feature maps by $\boldsymbol{f}(\boldsymbol{x}_i)$, respectively. Here $\boldsymbol{l}(\boldsymbol{x}_i) = [l_1(\boldsymbol{x}_i), \cdots, l_C (\boldsymbol{x}_i)  ]$, where $C$ represents the number of unique classes of skin images; $\boldsymbol{f}(\boldsymbol{x}_i) = [ f_1(\boldsymbol{x}_i), \cdots,  f_K(\boldsymbol{x}_i)] $ with $f_k(\boldsymbol{x}_i) \in \mathbb{R}^{H_f\times W_f} \ (k=1,\cdots, K)$, where $H_f, W_f$ denote the height and width of a feature map, respectively. 
     
\subsection{Logit-based dark knowledge}

\cite{hinton2015distilling} proposed a logit-based knowledge distillation method (a.k.a BLKD). Essentially, BLKD utilizes the (softened) logits of a teacher model as informative knowledge, which is then transferred to assist the training of a student model.    

BLKD includes two loss terms: 1) a term that matches the softened logits of a teacher model with those of a student one, and 2) a regular classification term that matches the predictions of a student model with the groundtruth label. For an image $\boldsymbol{x}_i$, denote its softened logits from a model as $\boldsymbol{p}(\boldsymbol{x}_i, T)$ at a temperature $T$, where $\boldsymbol{p}(\boldsymbol{x}_i, T) = [ p_1(\boldsymbol{x}_i, T), \cdots,  p_C(\boldsymbol{x}_i, T) ]$. Here $p_c(\boldsymbol{x}_i, T)$ denotes the soft probability that $\boldsymbol{x}_i$ belongs to a class $c$. With the softmax function, $p_c(\boldsymbol{x}_i, T)$ can be computed via the logit $l_c (\boldsymbol{x}_i)$,
\begin{equation}
    p_c(\boldsymbol{x}_i, T) = \frac{\textrm{exp}(l_c(\boldsymbol{x}_i) / T)}{\sum_{j=1}^C \textrm{exp} (l_j (\boldsymbol{x}_i) / T )}
    \label{eq:softmax}
\end{equation}
where a higher temperature $T$ provides softer probabilities for class predictions, and tends to indicate richer informative structures between classes. In BLKD, the loss of the first term measures the distribution differences of soft probabilities from a teacher (i.e., $\boldsymbol{p}^t(\boldsymbol{x}_i, T)$) and a student (i.e., $\boldsymbol{p}^s(\boldsymbol{x}_i, T)$) using the Kullback–Leibler (KL) divergence,

\begin{equation}
\begin{split}
        L_{KD} =  & \sum_{i=1}^B D_{KL} ( \boldsymbol{p}^t(\boldsymbol{x}_i , T) \ || \  \boldsymbol{p}^s(\boldsymbol{x}_i, T) ) \\
        = & \sum_{i=1}^B \sum_{c=1}^C p_c^t(\boldsymbol{x}_i, T) \ \textrm{log} \ \frac{p_c^t(\boldsymbol{x}_i, T)}{p_c^s(\boldsymbol{x}_i, T)}
\end{split}
\label{eq:kd_blkd}
\end{equation}
where $L_{KD}$ denotes the loss of the first term in BLKD that are computed on a batch of images. 

The predictions of a model can be obtained by simply setting $T=1$ in Eq.~(\ref{eq:softmax}). Then, as the second loss term in BLKD, a cross entropy loss can be computed between model predictions and groundtruth labels. Unlike general computer vision tasks, however, the second loss term is not  well suited for our skin cancer diagnosis problems. This is mainly due to the data imbalance issue in skin disease application, namely, we may only have few skin images for some diseases. Therefore, in this case, we turn to the weighted cross entropy (WCE). Denote by $w_c$ the weight for a class $c$, where $w_c$ can be computed as the inverse of the frequency for a disease class $c$ over the training dataset. Denote also the one-hot encoded label of $\boldsymbol{x}_i$ as $\boldsymbol{y}(\boldsymbol{x}_i) = [y_c(\boldsymbol{x}_i), \cdots, y_C(\boldsymbol{x}_i)]$.  Then, the WCE loss is expressed as, 
\begin{equation}
    L_{WCE} = \sum_{i=1}^B \sum_{c=1}^C - w_c \cdot y_c(\boldsymbol{x}_i) \ \textrm{log} \ p_c^s(\boldsymbol{x}_i, T=1)
\label{eq:wce}
\end{equation}

Then, the customized overall loss term of BLKD equals,
\begin{equation}
    L_{BLKD} = (1-\lambda_{kd}) \cdot L_{WCE} + \lambda_{kd} \cdot L_{KD} 
\label{eq:blkd}
\end{equation}
where $\lambda_{kd}$ is a hyperparameter to balance the two loss terms in BLKD.

\subsection{Inter-instance relational knowledge}

An inter-instance relational knowledge was proposed to model the structural relations among a batch of training data \citep{park2019relational}, termed data relational knowledge distillation (DRKD). Though there exist different variants of inter-instance knowledge (e.g., a kernelized form in \citep{peng2019correlation}), this study selects to use the popular DRLK due to its effectiveness and computational efficiency. To deal with the data scarcity challenge in skin cancer diagnosis, we also utilize such knowledge to additionally supervise the training of the student model.     

DRKD maps the embeddings of different instances (within a batch) into relational representations, then it minimizes the structural differences of such representations between the teacher and student models, such that the student can better mimic the teacher. Given instances $\{ \boldsymbol{x}_1, \cdots, \boldsymbol{x}_B \}$ and their embeddings denoted by $\{ \boldsymbol{e}_1, \cdots, \boldsymbol{e}_B \}$, \cite{park2019relational} defined two types of inter-instance relational knowledge: the distance-wise relation $\psi_d (\boldsymbol{e}_i, \boldsymbol{e}_j)$ between any two instances, and the angle-wise relation $\psi_a (\boldsymbol{e}_i, \boldsymbol{e}_j, \boldsymbol{e}_k)$ between any three instances. Here the potential functions $\psi_d (\boldsymbol{e}_i, \boldsymbol{e}_j)$ and $\psi_a (\boldsymbol{e}_i, \boldsymbol{e}_j, \boldsymbol{e}_k)$ are computed based on network embeddings $\boldsymbol{e}_i, \boldsymbol{e}_j, \boldsymbol{e}_k$ via the normalized Euclidean distance and angular distance, respectively. The loss term of DRKD is defined as,
\begin{equation}
\begin{split}
    L_{DRKD} &= \lambda_d \cdot \sum_{(\boldsymbol{x}_i, \boldsymbol{x}_j) \in \mathcal{X}_B^2} l(\psi_d^t(\boldsymbol{e}_i, \boldsymbol{e}_j), \psi_d^s(\boldsymbol{e}_i, \boldsymbol{e}_j)) \\
    &+ \lambda_a \cdot \sum_{(\boldsymbol{x}_i, \boldsymbol{x}_j, \boldsymbol{x}_k) \in \mathcal{X}_B^3} l(\psi_a^t(\boldsymbol{e}_i, \boldsymbol{e}_j, \boldsymbol{e}_k), \psi_a^s(\boldsymbol{e}_i, \boldsymbol{e}_j, \boldsymbol{e}_k))
\end{split}
\label{eq:drkd}
\end{equation}
where $\mathcal{X}_B$ represents a training subset comprised of a batch of instances, $l$ denotes the Huber function \citep{huber1992robust}; $\lambda_d, \lambda_a$ denote the weight for the distance-wise loss and the angle-wise loss \citep{park2019relational}, respectively.

\subsection{Intra-instance relational knowledge}

To further improve the performance of a student model, we propose to exploit the intra-instance relational knowledge using channel feature maps for each instance. We term this novel type of knowledge distillation method as channel relational knowledge distillation (CRKD). This work only extracts channel relations of feature maps at the last convolutional layer, since these feature maps contain more fine-grained and critical features for making final predictions. It is worth mentioning that the instance-level congruence term in \citep{peng2019correlation} actually refers to the KL divergence of logits between a teacher and a student (i.e., as in Eq.~\ref{eq:kd_blkd}), which is clearly different from our formulation.       

As illustrated in Figure~\ref{fig:flowchart}, given an input image $\boldsymbol{x}_i$, a (teacher or student) model extracts high-level features $f_k(\boldsymbol{x}_i)$ for $k=1,\cdots, K$. Then, CRKD can be performed following two sequential steps: the channel adaptation step, and the intra-instance relation generation step. Since the number of channels of a student model may not be equal to that of a teacher model, it is infeasible to align the intra-instance relations from two models directly. Therefore, for the student model, we adopt an adaptation module to match its channel number with that of the teacher. Specifically, we utilize a $1 \times 1$ convolution \citep{lin2014network} that linearly projects the channel feature maps of the student model to a space that shares a same channel number with the teacher model \citep{romero2015fitnets}.   

In the second step, we will model the relation between an arbitrary (vectorized) feature map pair of an instance $\boldsymbol{x}_i$ as $\big(\textrm{Vec}(f_k(\boldsymbol{x}_i)), \textrm{Vec}(f_{k'}(\boldsymbol{x}_i)) \big)$ utilizing the un-normalized cosine similarity metric, 
\begin{equation}
    r\big(f_k(\boldsymbol{x}_i), f_{k'}(\boldsymbol{x}_i)\big) = \big<\textrm{Vec}(f_k(\boldsymbol{x}_i), \textrm{Vec}(f_{k'}(\boldsymbol{x}_i) \big>
\end{equation}
Denote the overall intra-instance relational matrix as $\boldsymbol{R} \in \mathbb{R}^{K\times K}$, then $\boldsymbol{R}$ can be generated with entries $\boldsymbol{R}_{k, k'}$ as,
\begin{equation}
   \boldsymbol{R}_{k, k'} (\boldsymbol{x}_i) = r\big(f_k(\boldsymbol{x}_i), f_{k'}(\boldsymbol{x}_i)\big), \ k, k' \in \{ 1, \cdots, K\}
\end{equation}
A more computationally efficient way is adopt the Grammian matrix \citep{johnson2016perceptual,ding2021towards} to generate this relational matrix. To encourage the student learn the intra-instance relations of the teacher, we define the CRKD loss term over a batch of samples,
\begin{equation}
    L_{CRKD} = \sum_{i=1}^B \frac{1}{K H_f W_f} \big|\big| \boldsymbol{R}^t (\boldsymbol{x}_i) - \boldsymbol{R}^s (\boldsymbol{x}_i) \big|\big|_F 
\end{equation}
where $F$ denotes the Frobenius norm of matrices; $H_f, W_f$ refer to the height and width of a feature map of the teacher model, respectively.

The intra-instance relation formulation above appears to resemble the angle-wise relation in DRKD, but here we are considering a two-tuple relation scenario. Another major distinction is that we are modeling the relation between any two channel feature maps of one instance rather than the embeddings from different instances. Our CRKD also differs from \citep{yim2017gift} and \citep{liu2019knowledge} that aim to measure the correlations across different network layers which tend to involve considerably more computational costs and instability issues. Besides, without the channel adaptation module as in our method, they require a similar network architecture (e.g., from a same architecture family) for the teacher and student models. This strict requirement, however, restrains their applicability in our considered scenario to establish more portable skin lesion diagnosis models. 

A physical interpretation of our intra-instance relational modeling is to extract the texture information of an input image $\boldsymbol{x}_i$ in the feature space. The extracted textures have been demonstrated crucial in computer vision tasks, such as image recognition \citep{geirhos2018imagenet} and visual tracking attacks \citep{ding2021towards}. Since textural features also contain important discriminative knowledge in the skin lesion diagnosis task, this type of relation and knowledge can provide auxiliary favorable cues to facilitate the training of our student model.

\subsection{Diverse knowledge integration}

To further boost the student training process, we integrate diverse levels of knowledge and transfer them to the student model. As depicted in Figure~\ref{fig:flowchart}, our diverse knowledge distillation (D-KD) loss can be expressed as,
\begin{equation}
    L_{D-KD} = \lambda_{blkd} \cdot L_{BLKD} + \lambda_{drkd} \cdot L_{DRKD} + \lambda_{crkd} \cdot L_{CRKD}
\label{eq:overall}
\end{equation}
where $\lambda_{blkd}, \lambda_{drkd}, \lambda_{crkd}$ denote the hyperparameter to balance the loss terms for BLKD, DRKD, and CRKD, respectively. 

In addition, we can incorporate the self-supervised training strategy \citep{xu2020knowledge} into the diverse knowledge-integrated distillation loss $L_{D-KD}$ to additionally improve the training dynamics of the student model. Self-supervised learning (SSL) is an effective strategy to learn generic visual representations from unlabeled data for downstream tasks (e.g., image classification, semantic segmentation) in the computer vision field \citep{chen2020simple,he2020momentum,jing2020self}. Moreover, a very recent work \citep{liu2022self} shows that this learning strategy can well alleviate the data imbalance issue, which further validated our motivation to incorporate the self-supervision knowledge. 

A typical SSL scheme is to employ a contrastive loss \citep{van2018representation} that maximizes the agreement in feature space of positive sample pairs generated from different augmentation transformations on a same image \citep{he2020momentum,jing2020self}. For the first time, Xu et al. \citep{xu2020knowledge} introduced SSL into knowledge distillation (SSKD), and conducted a thorough study on the effectiveness of this learning strategy in the KD setting. To perform the self-supervision, SSKD pre-trains a teacher model with the contrastive loss, then transfers the self-supervised predictions from the teacher to the student. Following \citep{xu2020knowledge}, we integrate the self-supervised training strategy into our D-KD framework. Then, the loss function for the self-supervised diverse knowledge distillation (SSD-KD) is,
\begin{equation}
    L_{SSD-KD} = L_{D-KD} + \lambda_{sskd} \cdot L_{SSKD} 
\label{eq:ssd-kd}
\end{equation}
where $\lambda_{sskd}$ denotes the hyperparameter for the self-supervision regularizer. 

In Algorithm ~\ref{alg}, we describe the details of the proposed method.

\begin{algorithm}[tb]
	\footnotesize
	\SetAlgoLined
	\KwData{A training dataset $(\mathcal{X}, \mathcal{Y})$, a pretrained teacher model, an untrained student model, batch size $B$, the maximum number of training epochs $N$, hyperparameters $\lambda_{blkd}, \lambda_{drkd}, \lambda_{crkd}, \lambda_{sskd}$, a minibatch stochastic gradient descent optimizer, an early stop epoch $K_{stop}$.} 
	Initialize the student model using random initialization, set the early stopping flag as False\;
	Split $(\mathcal{X}, \mathcal{Y})$ into $\ceil{N/B}$ non-overlapping batches randomly\;
	\For{$n=1$ \KwTo $N$}{
	Reset the knowledge distillation loss $L$ as 0\;
	\For{$iter=1$ \KwTo $\ceil{N/B}$}{
	Draw a batch of skin images $\{ \boldsymbol{x}_1, \cdots, \boldsymbol{x}_B \}$\;
	Input the image batch into the teacher model, computes intra, inter-instance relational knowledge, and logit-based knowledge\;
	Input the image batch into the student model, computes intra, inter-instance relational knowledge, and logit-based knowledge\;
	Compute the knowledge distillation loss using Eq. (\ref{eq:ssd-kd}) for SSD-KD\;
	Update the student model using the optimizer\;
	}
	If loss not decrease within $K_{stop}$ epochs: stop the program\;
	Else: continue the program\;
	}
	\textbf{Return}: An optimized portable student model.
	\caption{The algorithm of our method.}
	\label{alg}
\end{algorithm}

\section{Experiments and results}
In this section, we conduct experiments to empirically demonstrate the effectiveness of the proposed method. We will describe in detail our experimental setup, evaluation metrics, and then present the comparison results and our analysis.  

\subsection{Dataset}

\begin{figure*}
\centering
\includegraphics[width=1\textwidth]{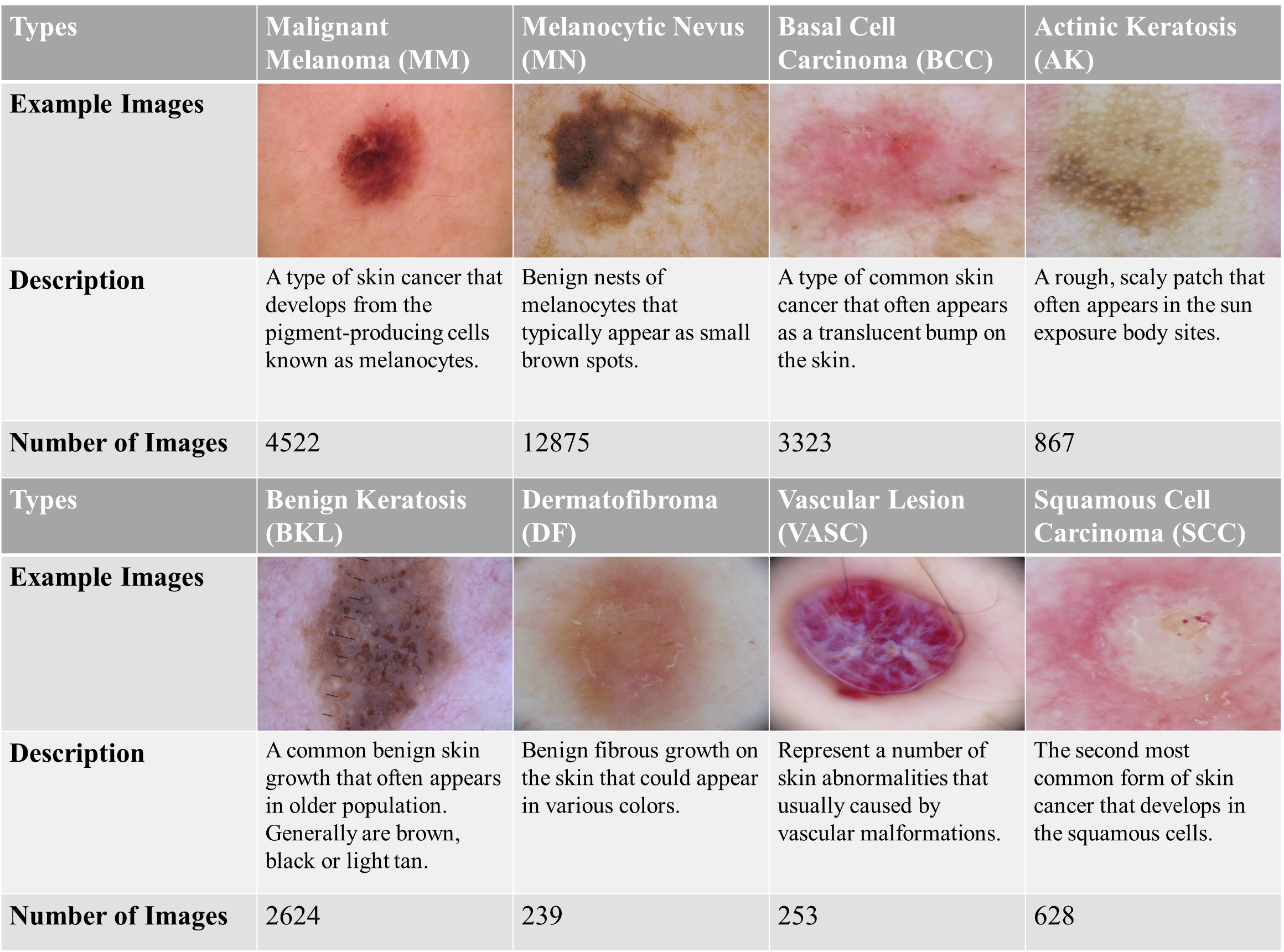}
\caption{\label{fig:data}Example images along with their detailed information from the ISIC 2019 dataset.}
\end{figure*}

The dataset we used in this study is ISIC 2019 \citep{tschandl2018ham10000,codella2018skin,combalia2019bcn20000}, the large-scale publicly available dermoscopy imaging database, that contains 25331 dermoscopic images acquired from different resources along with their diagnosis labels as their ground truth. Part of the images also have corresponding meta information, such as gender, age, anatomy site, etc. The ISIC 2019 contains 8 different skin diseases including malignant melanoma (MM), melanocytic nevus (MN), basal cell carcinoma (BCC), actinic keratosis (AK), benign keratosis (BKL), dermatofibroma (DF), vascular lesion (VASC) and squamous cell carcinoma (SCC). An overview of details for each diseases along with the exampled images is presented in  Figure~\ref{fig:data}.

\subsection{Implementation details}

In the experiment part, we selected ResNet50 as the teacher model while MobileNetV2 was chosen as the student model. ResNet50 is one of the most powerful convolutional neural network (CNN) architectures and has demonstrated outstanding representation and generalization abilities in medical imaging analytic area \citep{raghu2019transfusion, zhang2019medical, xie2021survey}. However, ResNet50 increases the complexity of architecture and requires huge computational power during training. Therefore, ResNet50 is a suitable choice for teacher model in our project. On the other hand, MobileNetV2, as an advanced version of MobileNet, benefits from its reduced network size and number of parameters, and it has been proven as an effective feature extractor for object detection for mobile applications \citep{sandler2018mobilenetv2, tougaccar2021intelligent, srinivasu2021classification}. In this project, the carefully selected teacher model and student model not only have huge differences in model size, but also process of dramatic distinctions in the backbone structure. Unlike other knowledge distillation studies that only focus on the same structure with different number of layers, our task is more challenging and this setup can better validate the practicality and applicability of the KD methods in the field of dermoscopy imaging.

We implemented our proposed SSD-KD method using PyTorch \citep{paszke2019pytorch} on one NVIDIA Tesla V100 GPU card (32 GB memory) card at the UBC ARC Sockeye, which is a high-performance computing platform available to UBC researchers across all disciplines. For a total of 25331 images from the ISIC2019, following general practices, we randomly selected 10\% of the images to compose a test set, while the rest images were also randomly divided into training and validation with a ratio of 8: 2. Before feeding images into the deep learning models, we uniformly resized all the images to $224 \times 224$ and applied data augmentation using common image processing operations, including horizontal and vertical flips, adjustments in brightness, contrast, saturation, image scaling, and random noise. The teacher and student models were pre-trained on ImageNet \citep{deng2009imagenet} with the weights gathered from torchivision API and then fine-tuned for 150 epochs using the mini-batch SGD optimizer with an initial learning rate of 0.001, momentum of 0.9, and the weight decay of 0.001. For hyperparameters in D-KD and SSD-KD, we used default values for parameters within modules BLKD \citep{hinton2015distilling}, DRKD \citep{park2019relational}, and SSKD \citep{xu2020knowledge}. Then for D-KD, we set $\lambda_{blkd}, \lambda_{drkd}, \lambda_{crkd}$ as 1, 1, and 1000, respectively. The parameters were tuned on the validation dataset with grid search. In SSD-KD, since the SSKD module already included BLKD in it \citep{xu2020knowledge}, we therefore treated SSKD as a whole module and used their default parameters. We then set $\lambda_{drkd}, \lambda_{crkd}, \lambda_{sskd}$ as 1, 1000 and 1, respectively.   

In the training phase, we used a conditional reduction strategy for the adjustment of the learning rate. If the model did not improve for 10 consecutive epochs, the learning rate value was reduced by a rate of 0.1. Early stopping was also used for every 15 consecutive epochs if the model did not improve. In the initial experiments, we found that increasing the batch size within a certain range will significantly improve the learning ability of the model. Therefore, while considering the computational cost, on the premise of obtaining a better teacher model, we set the batch size as 128. Additionally, because of the severe data imbalance problems between different skin diseases, we used the weighted cross-entropy as the loss function. The weights are determined according to the frequency of diagnosis labels and thus assigning more attention to the classes that less appeared in the training dataset.

\subsection{Evaluation metrics}

We employed four widely used metrics to evaluate the prediction performance for multi-classes skin diseases of our proposed method. The metrics include Accuracy (ACC), Balanced Accuracy (BACC), mean Average Precision (mAP), and Area Under the Curve (AUC).

\begin{table*}[width=.9\textwidth,cols=4,pos=h]

 \caption{The comparison of the prediction accuracy and computation costs on ISIC 2019 between the teacher model(ResNet50) and the student model (MobielNetv2).}\label{tbl1}
 \begin{tabular*}{\tblwidth}{@{}LLLLLLL@{} }
\toprule
 Model & Input Size & ACC & Model Size & Number of parameters & Time per inference step & GFlops\\  
\midrule
ResNet50 & $224\times224$ & 0.820 & 98MB & 25.6M& 58.2ms & 4.11\\
MobileNetV2 & $224\times224$ & 0.754 & 16MB &3.5M & 22.9ms & 0.31\\
\bottomrule
\end{tabular*}
\end{table*}
\begin{table*}[width=.9\textwidth,cols=4,pos=h]

 \caption{Comparison of single teacher/models and other state-of-the-art knowledge distillation methods. Best results are highlighted in bold for each column.}
 \label{tbl2}
 \begin{tabular*}{\tblwidth}{@{}LLLLLLL@{} }
\toprule
Teacher & Student & KD Method & ACC & BACC & AUC& mAP\\     
\midrule
ResNet50 & N/A & N/A & 0.820 &0.822 & 0.975 & 0.740 \\   
MobileNetV2 & N/A & N/A & 0.754 & 0.767 & 0.959 & 0.625 \\ 
ResNet50 & MobileNetV2 & BLKD \citep{hinton2015distilling} & 0.838 &0.824 & 0.976 & 0.751 \\  
ResNet50 & MobileNetV2 & FitNet \citep{romero2015fitnets} & 0.807 &0.806 & 0.971 & 0.700 \\ 
ResNet50 & MobileNetV2 & DRKD \citep{park2019relational} & 0.816 &0.804 & 0.972 & 0.744 \\ 
ResNet50 & MobileNetV2 & SSKD \citep{xu2020knowledge} & 0.840 &0.840 & \textbf{0.978} & 0.759 \\ 
ResNet50 & MobileNetV2 & SSKD+DRKD+CRKD (ours) & \textbf{0.846} &\textbf{0.843} & 0.977& \textbf{0.796 }\\ 
\bottomrule
\end{tabular*}
\end{table*}

\begin{table*}[width=.9\textwidth,cols=4,pos=h]
\caption{Comparison of different modules equipped in SSD-KD for ablation studies. Best results are highlighted in bold for each column.}
\label{tbl3}
\begin{tabular*}{\tblwidth}{@{}LLLLLLL@{} }
\toprule
Teacher & Student & KD Method & ACC & BACC & AUC& mAP\\     
\midrule
ResNet50 & MobileNetV2 & BLKD+DRKD & 0.839 &0.827 & 0.977 & 0.755 \\ 
ResNet50 & MobileNetV2 & BLKD+CRKD & 0.839 &0.824 & 0.977 & 0.773 \\ 
ResNet50 & MobileNetV2 & SSKD+DRKD & 0.841 &0.840 & 0.978 & 0.769 \\ 
ResNet50 & MobileNetV2 & SSKD+CRKD & 0.842 &0.839 & 0.977 & 0.773 \\ 
ResNet50 & MobileNetV2 & BLKD+DRKD+CRKD (D-KD) & 0.844 &0.839 & \textbf{0.978} & 0.788 \\ 
ResNet50 & MobileNetV2 & SSKD+DRKD+CRKD (SSD-KD) & \textbf{0.846} &\textbf{0.843} & 0.977 & \textbf{0.796} \\ 
\bottomrule
\end{tabular*}
\end{table*}

\subsection{Comparison of the teacher and student models}

To clearly indicate the differences between the teacher model and the student model in various indicators, we compared the classification accuracy along with the computational costs in detail. In Table~\ref{tbl1}, we measured the computational costs by four commonly used complexity standards, including model size, number of parameters, time for every inference step, and the flops. Both teacher model and student model were trained without any knowledge of distillation. As shown in Table~\ref{tbl1}, the teacher model achieved much better performance in skin lesion classification with an accuracy of 82.0\% , while the lightweight student model only got 75.4\% in the same task. However, the model size of the teacher model reached 98M and the number of parameters was 25.6M, which was over 6 times that of the student model. In addition, the time for each inference step of the teacher model was 58.2ms, while the student model only used half of this time. The flops, which were used to measure the number of operations of the network, indicated that the teacher model owned over 10 times of complexity in terms of the operation numbers. The above data comparison showed that the teacher model can indeed achieve an essential improvement in the model's predictive ability, but it is also accompanied by a greater computational requirement. This is well aligned with our assumptions for this project.

\subsection{Comparison with other KD methods}

\begin{figure*}[h]
  \centering
  
  \subfigure[MobileNetV2]{
  \includegraphics[width = 5cm]{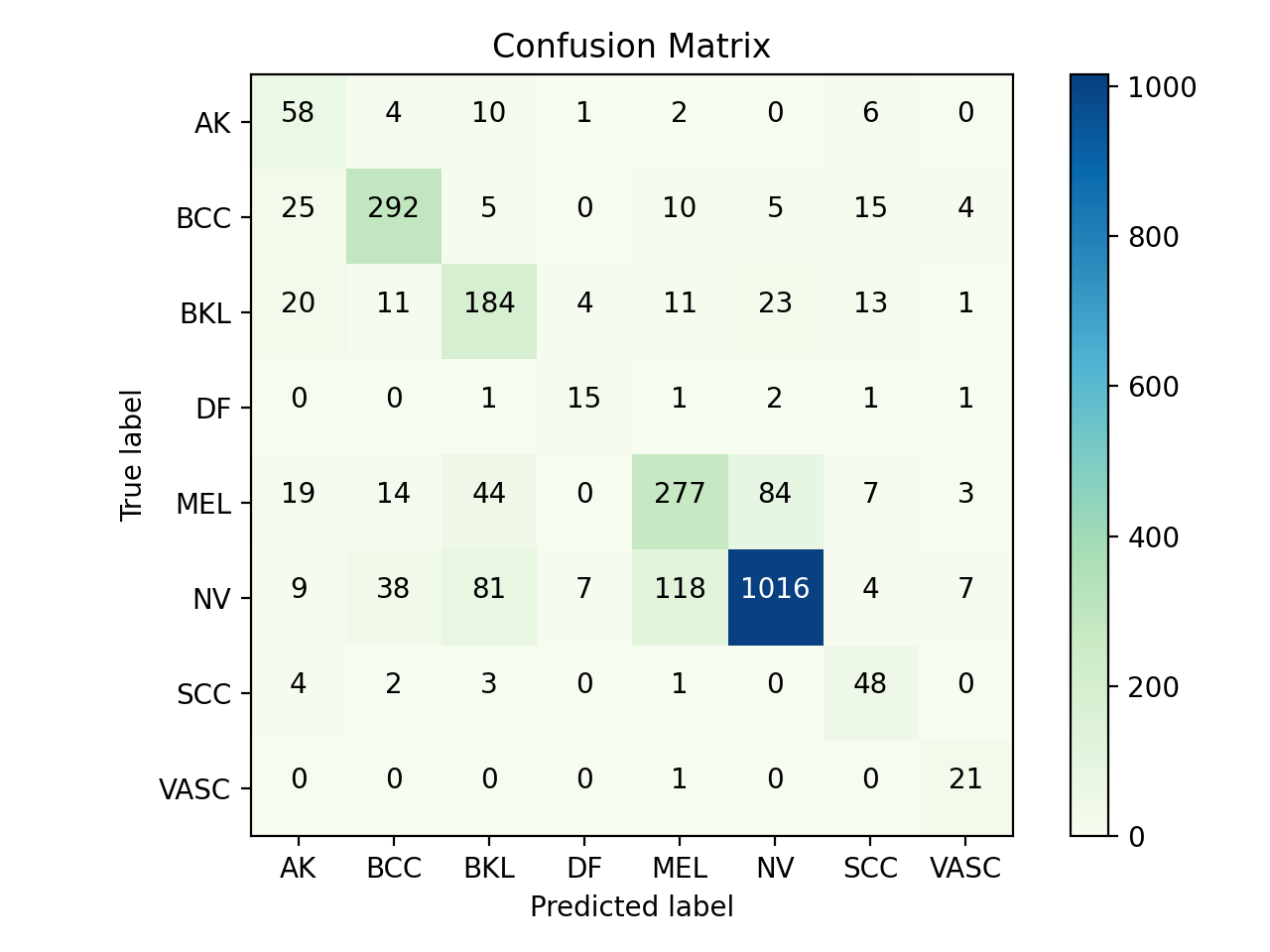}
  }
  \quad
  \subfigure[ResNet50]{
  \includegraphics[width = 5cm]{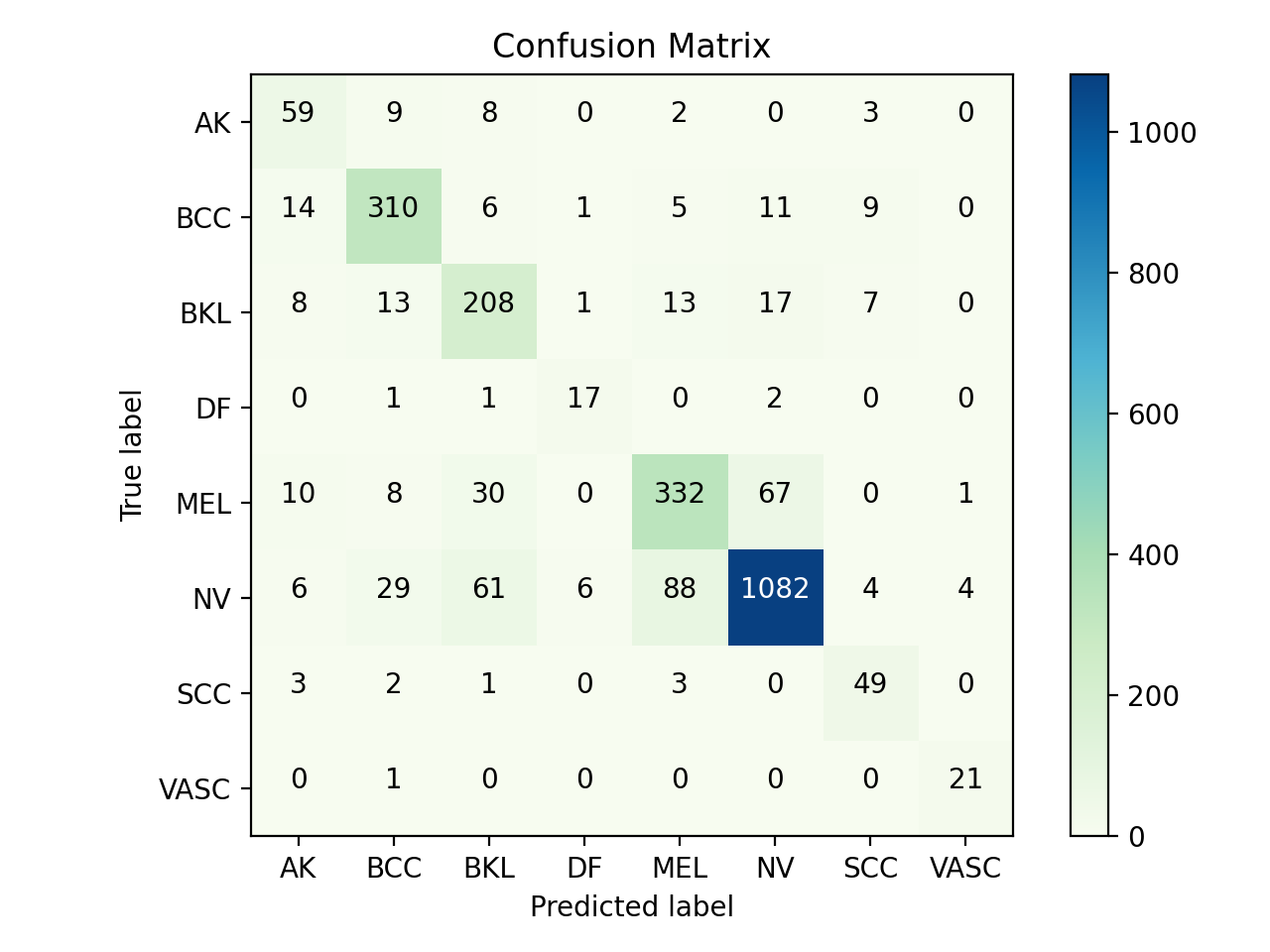}
  }
  \quad
  \subfigure[BLKD]{
  \includegraphics[width = 5cm]{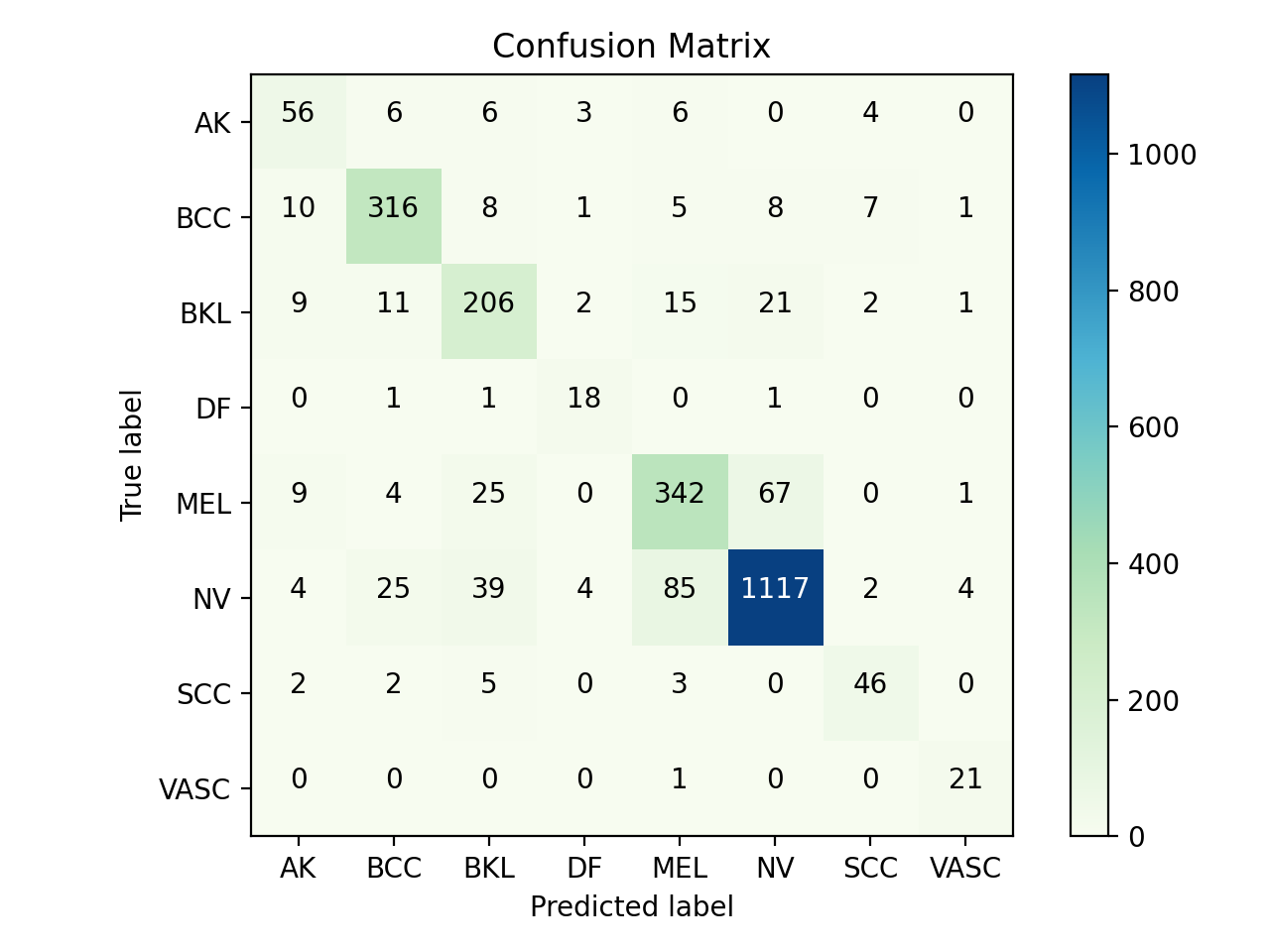}
  }
  \quad
  \subfigure[FitNet]{
  \includegraphics[width = 5cm]{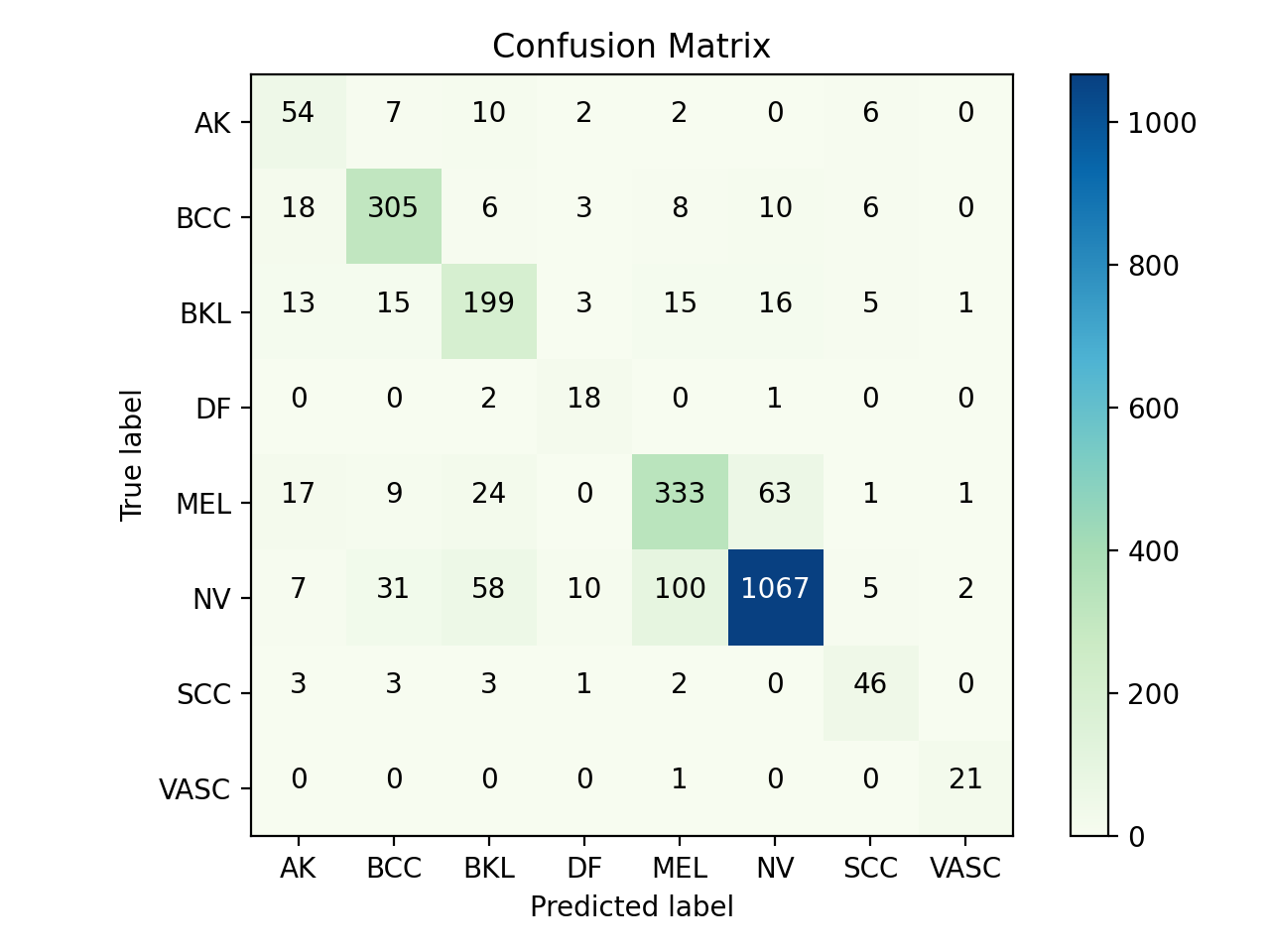}
  }
  \quad
  \subfigure[DRKD]{
  \includegraphics[width = 5cm]{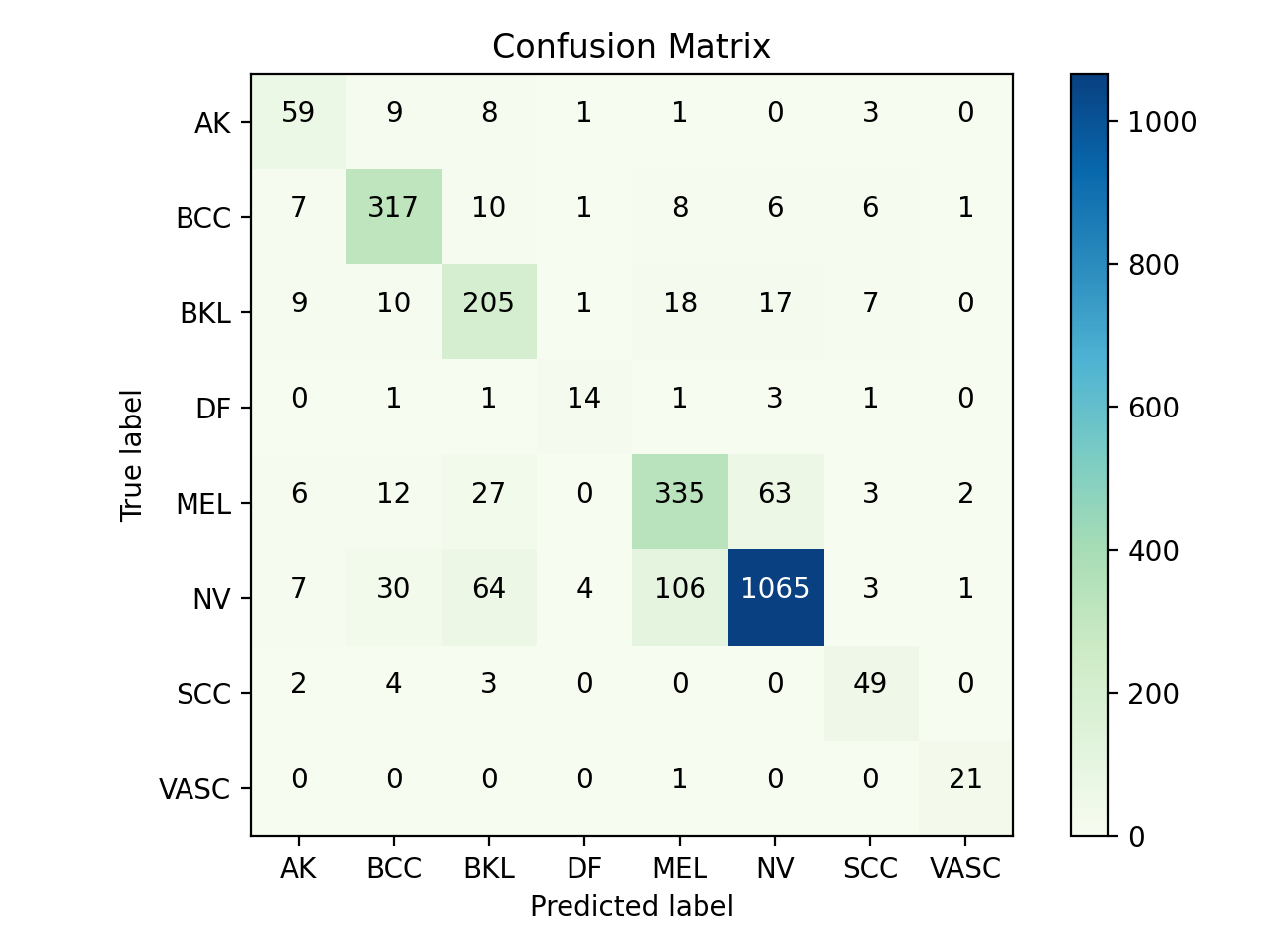}
  }
  \quad
  \subfigure[SSKD]{
  \includegraphics[width = 5cm]{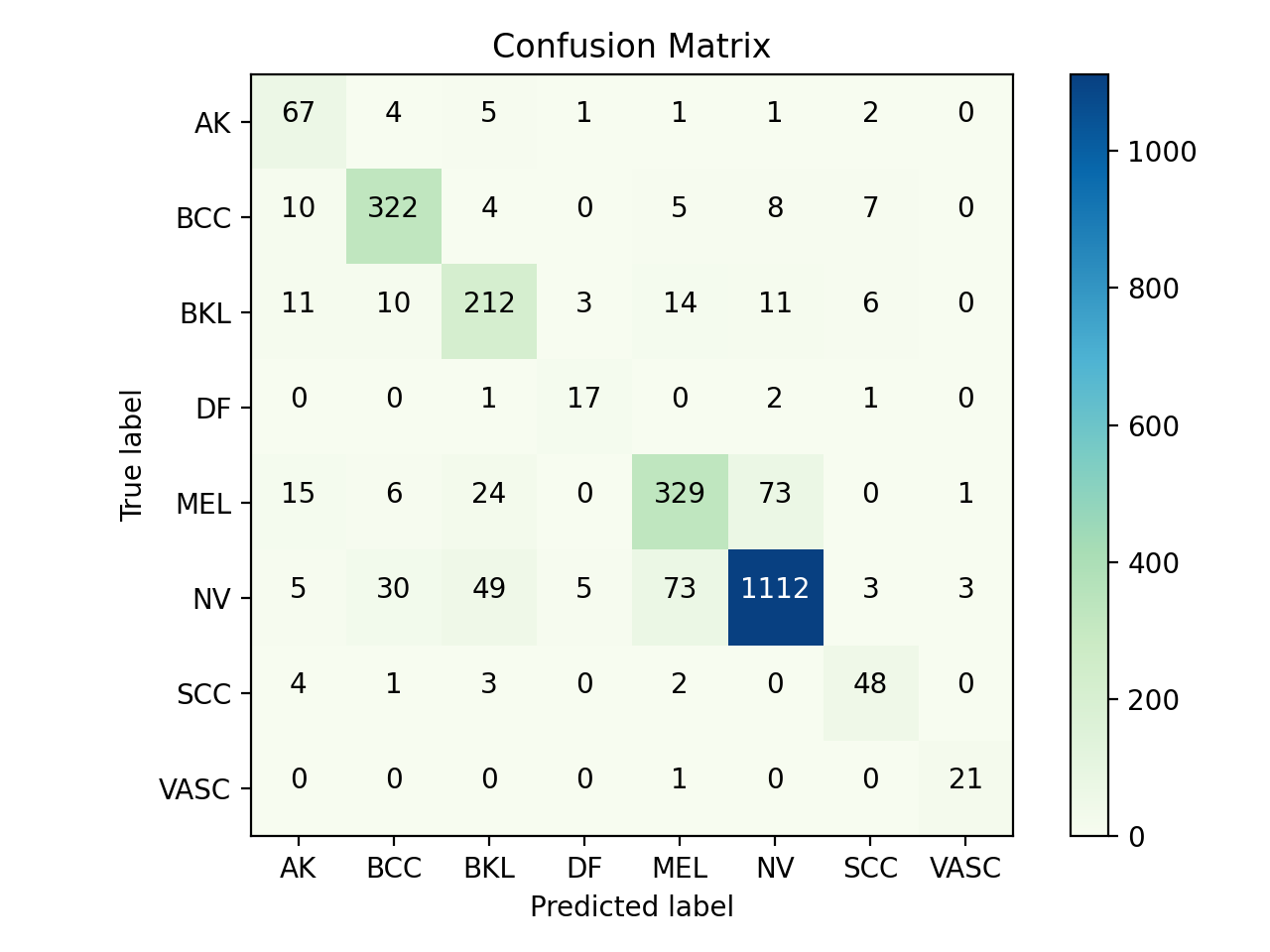}
  }
  \quad
  \subfigure[BLKD+DRKD]{
  \includegraphics[width = 5cm]{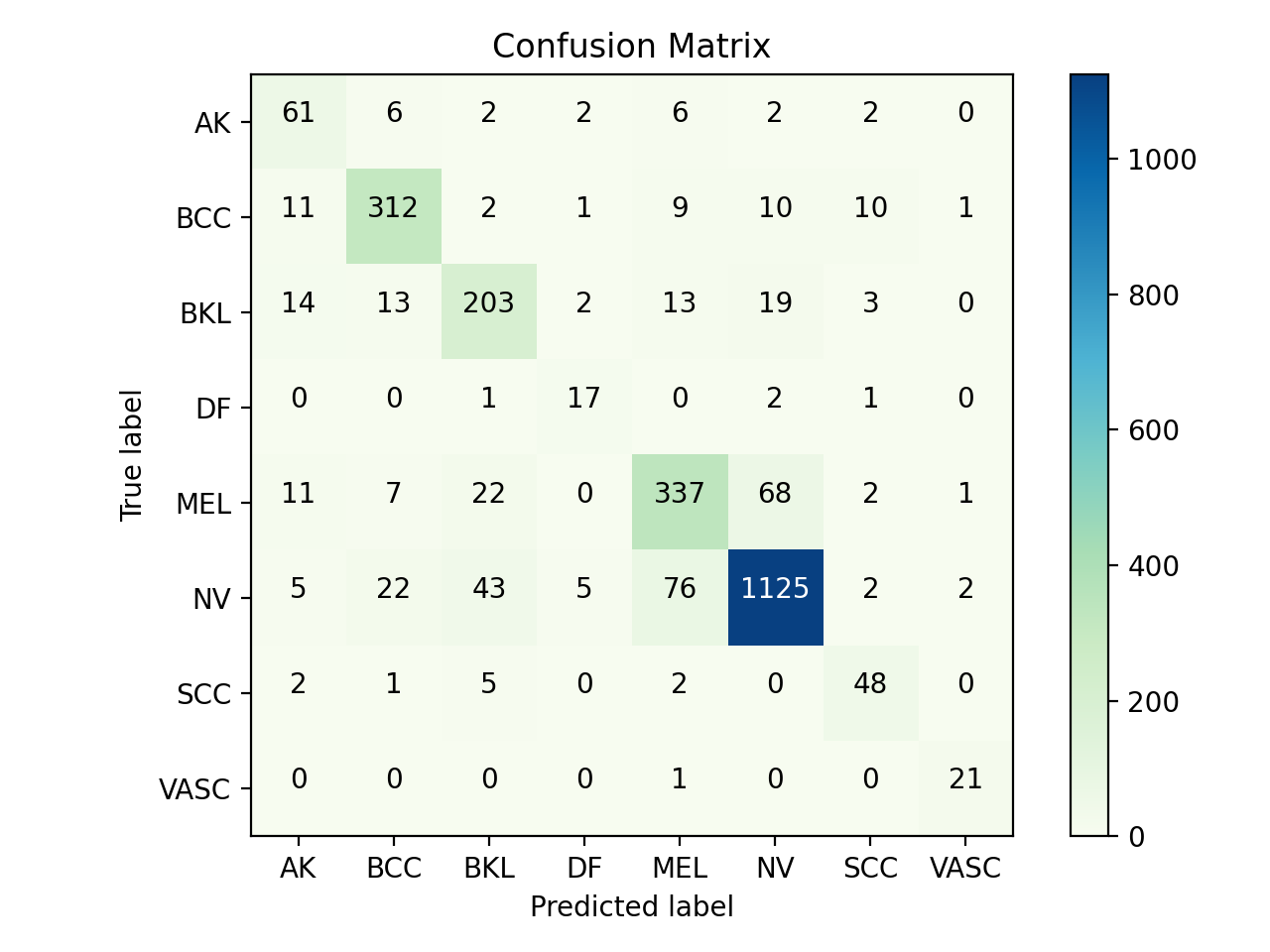}
  }
  \quad
  \subfigure[BLKD+CRKD]{
  \includegraphics[width = 5cm]{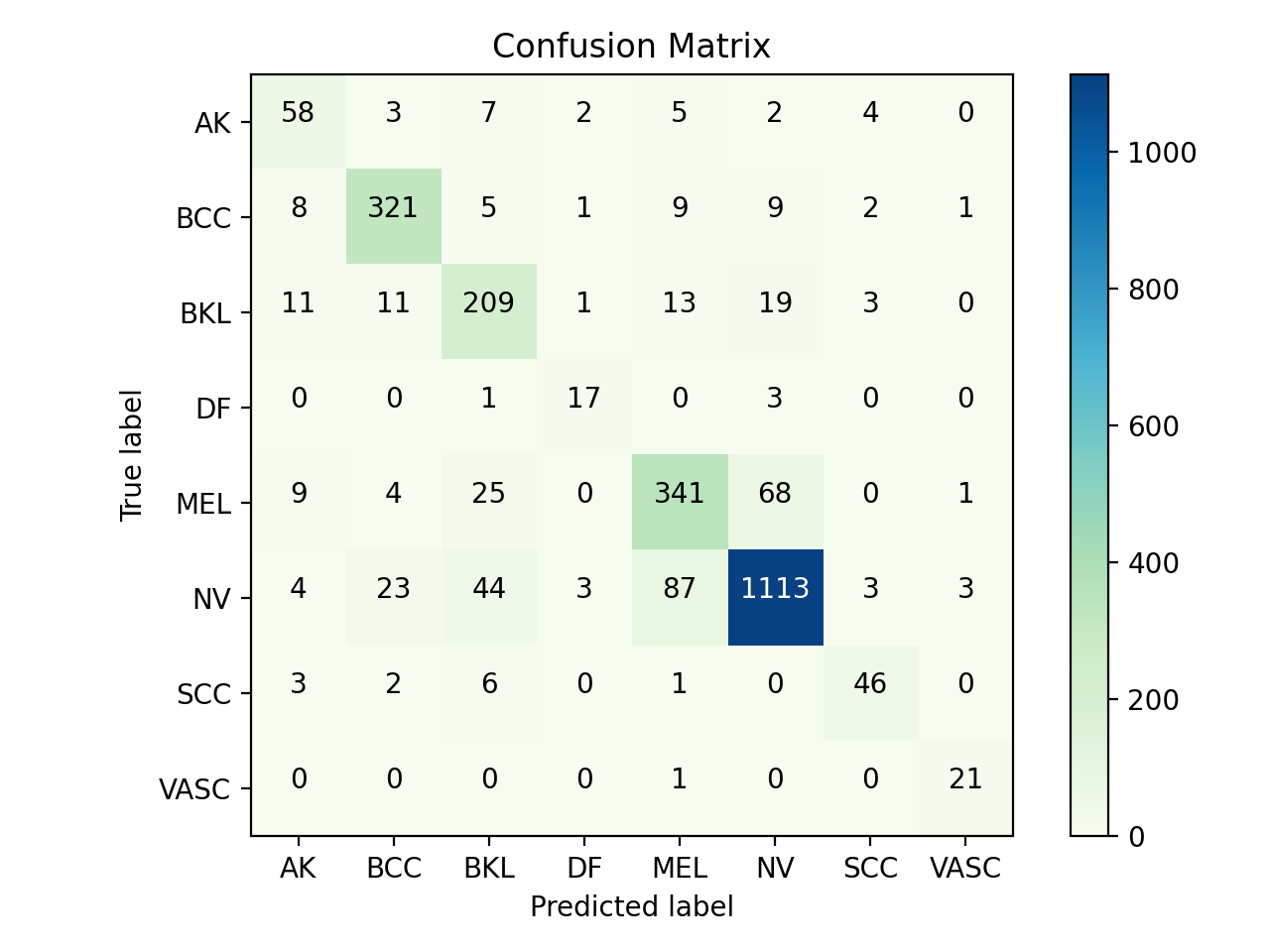}
  }
  \quad
  \subfigure[SSKD+DRKD]{
  \includegraphics[width = 5cm]{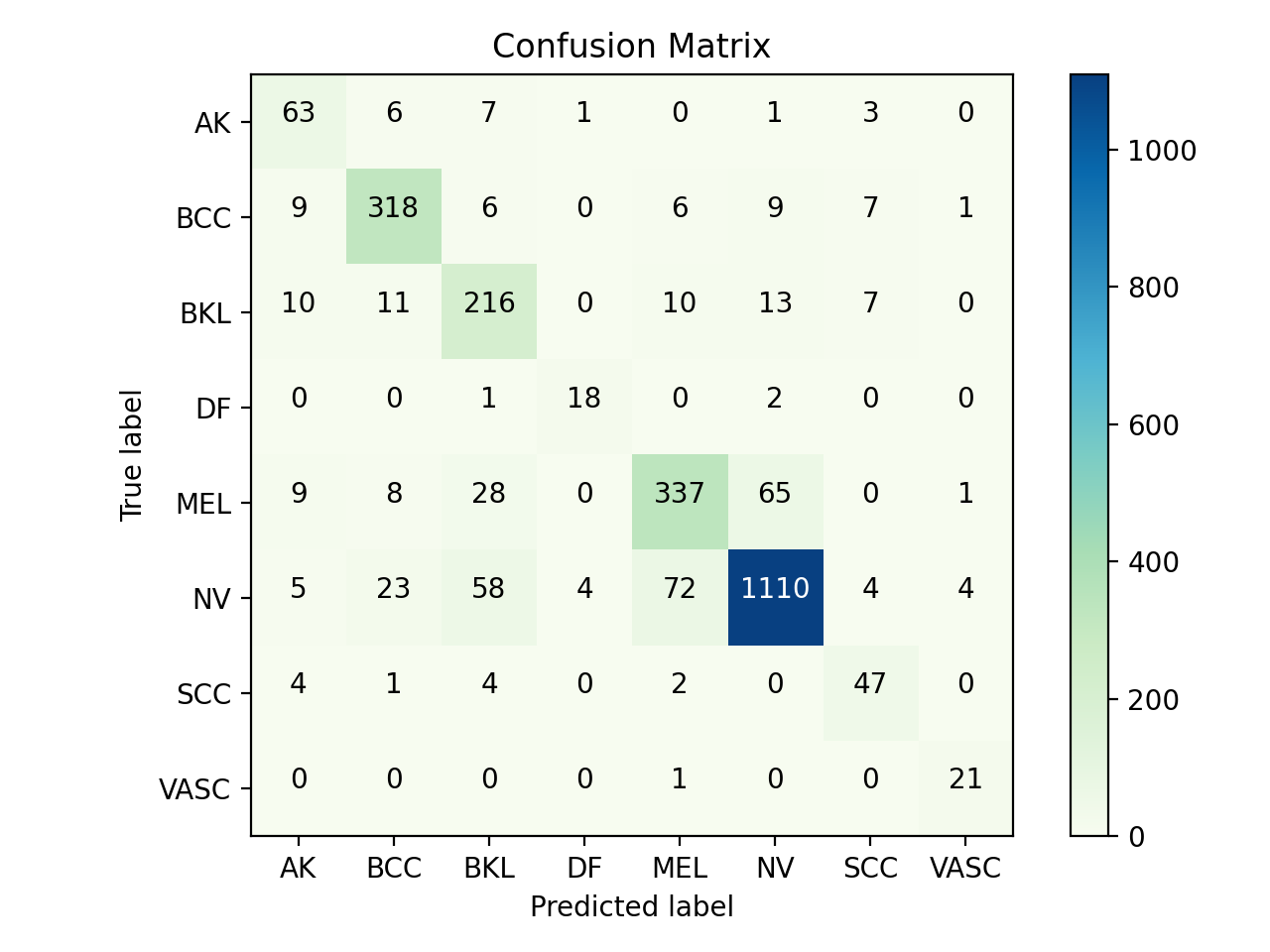}
  }
  \quad
  \subfigure[SSKD+CRKD]{
  \includegraphics[width = 5cm]{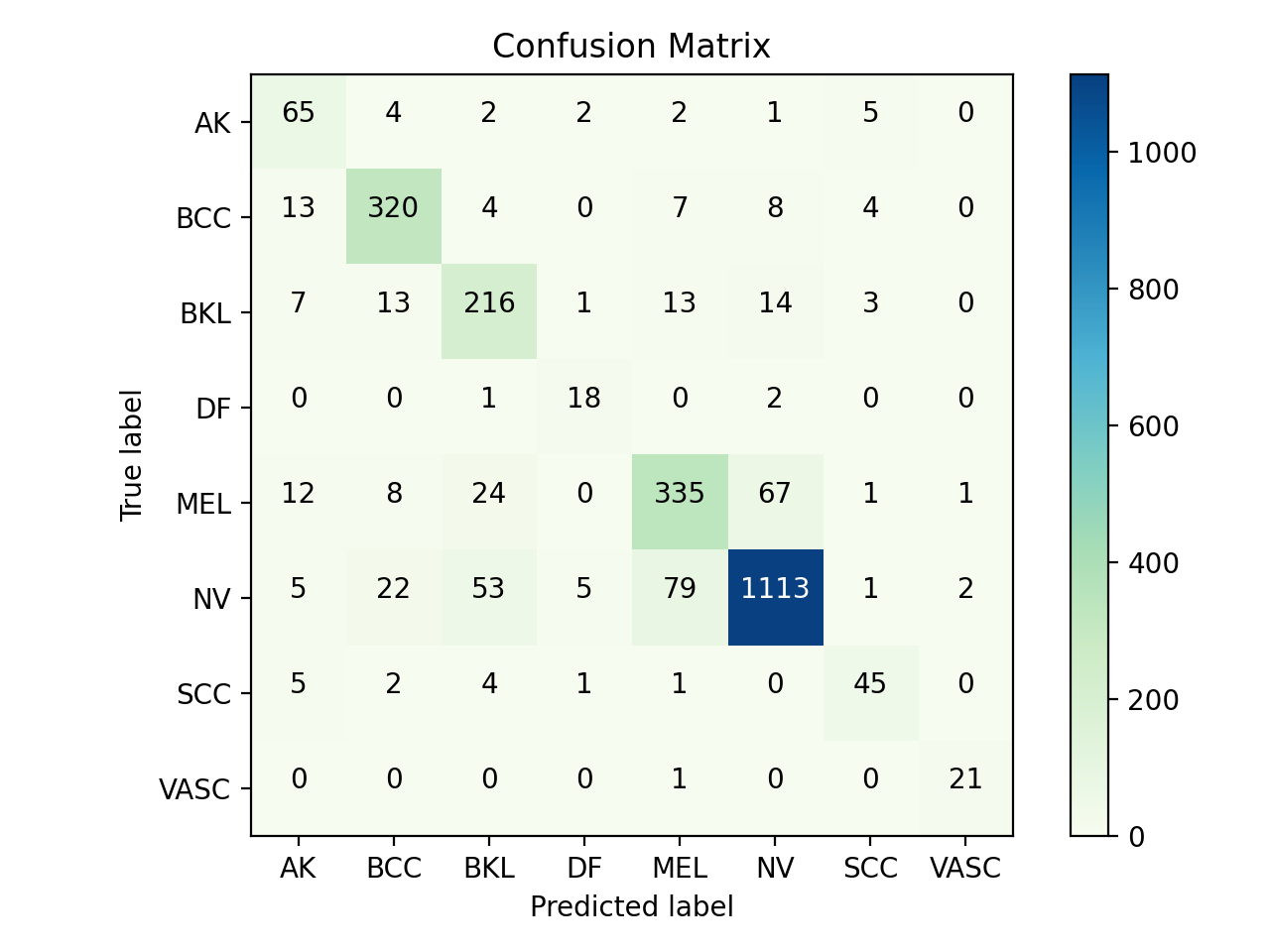}
  }
  \quad
  \subfigure[BLKD+DRKD+CRKD]{
  \includegraphics[width = 5cm]{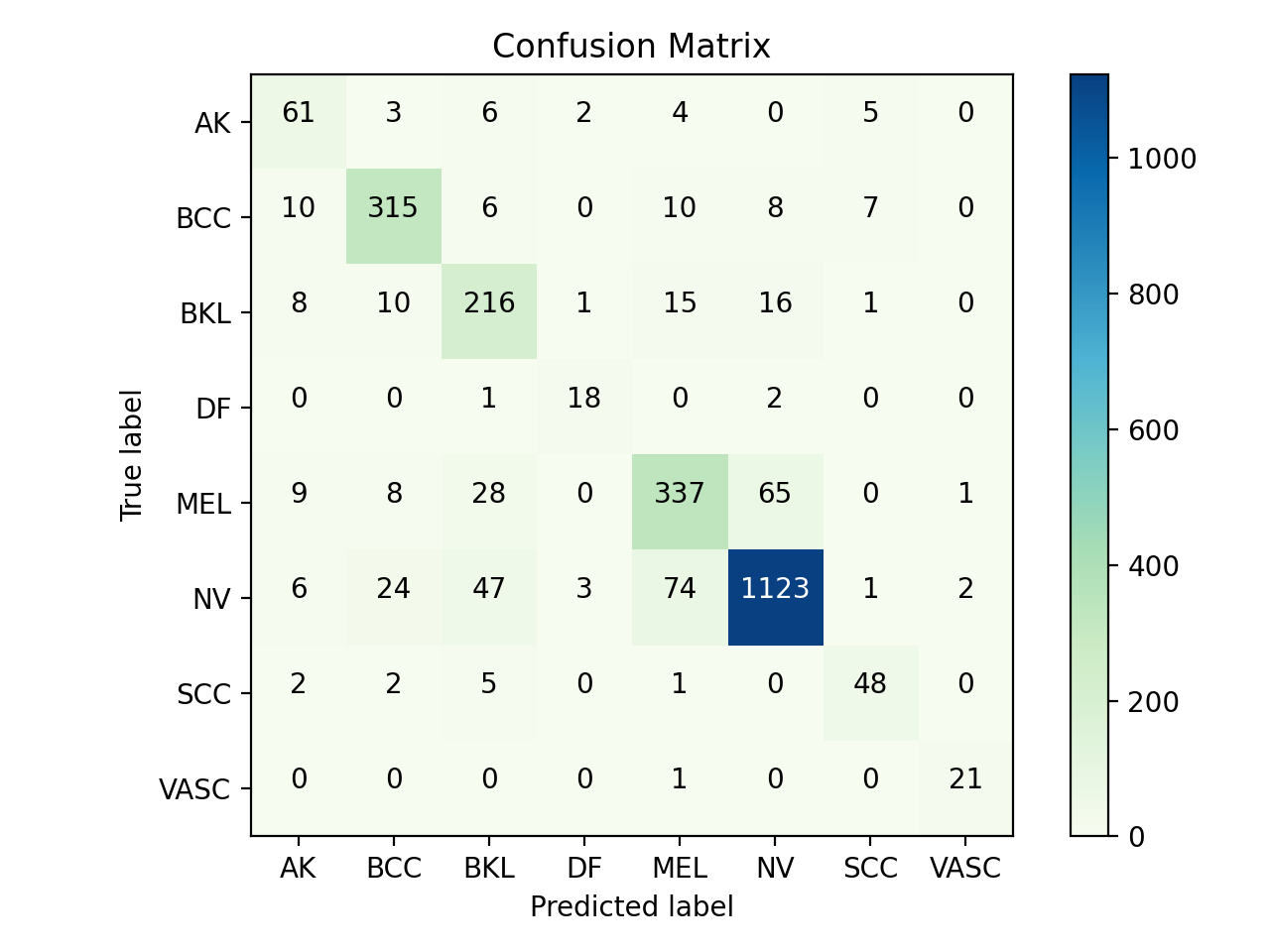}
  }
  \quad
  \subfigure[SSKD+DRKD+CRKD]{
  \includegraphics[width = 5cm]{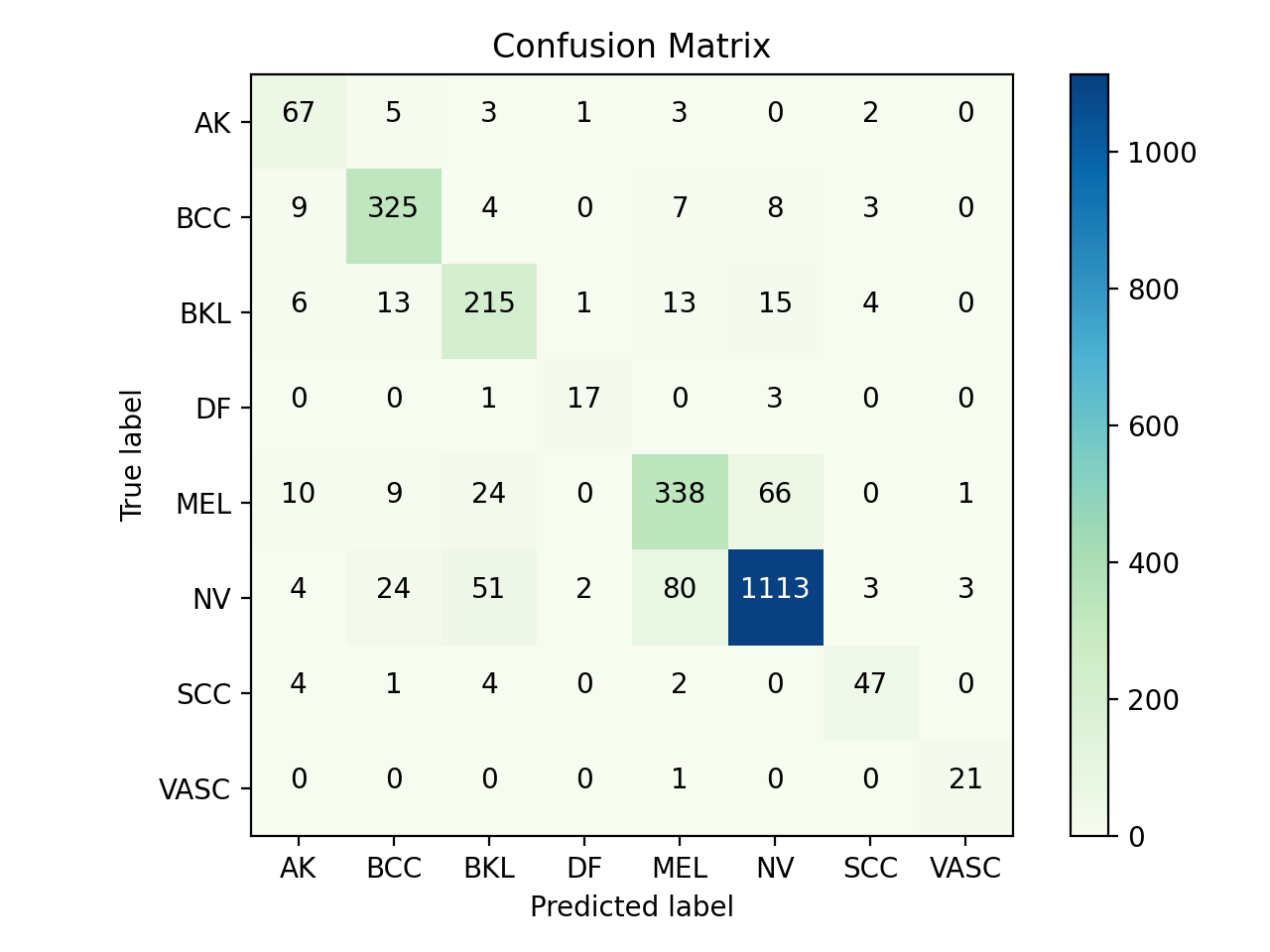}
  }
  \caption{Visualizations of confusion matrices from skin lesion classification models that are trained: without KD (as shown in subfigures (a) -- (b)), and with different KDs (as shown in subfigures (c) -- (l)).}
  \label{fig:conf_mat}
\end{figure*}

\begin{figure*}[htbp]
  \centering
  
  \subfigure[MobileNetV2]{
  \includegraphics[width = 5cm]{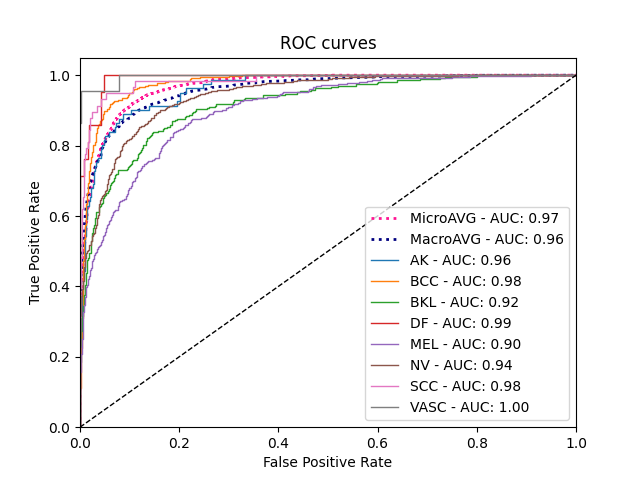}
  }
  \quad
  \subfigure[ResNet50]{
  \includegraphics[width = 5cm]{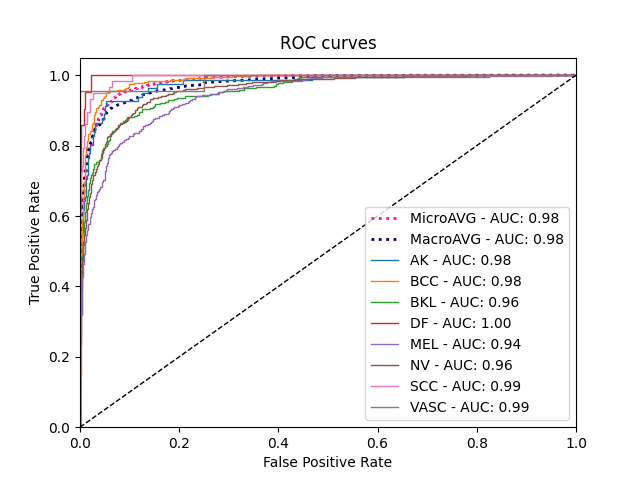}
  }
  \quad
  \subfigure[BLKD]{
  \includegraphics[width = 5cm]{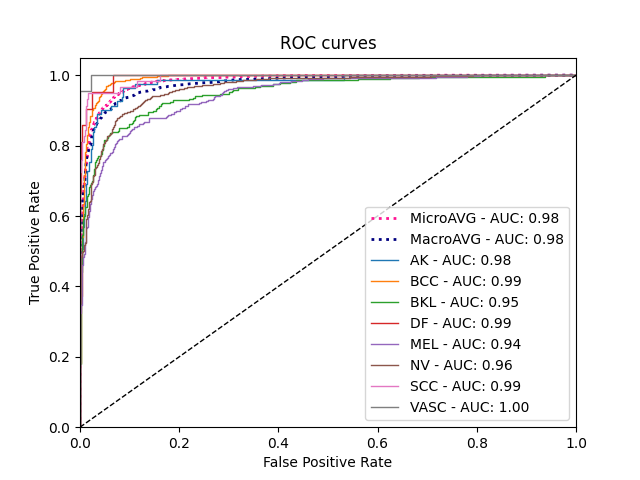}
  }
  \quad
  \subfigure[FitNet]{
  \includegraphics[width = 5cm]{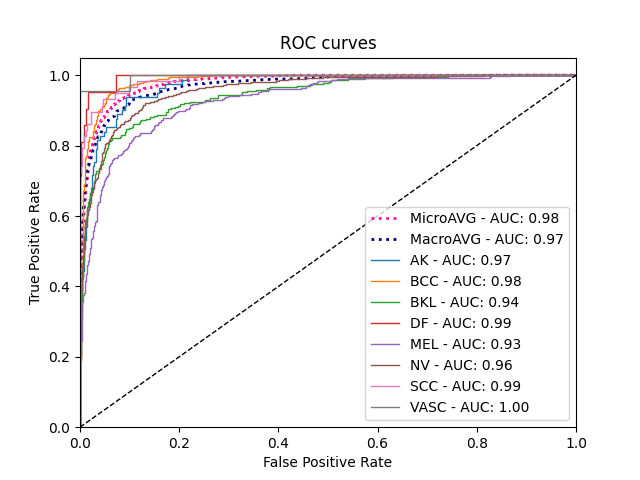}
  }
  \quad
  \subfigure[DRKD]{
  \includegraphics[width = 5cm]{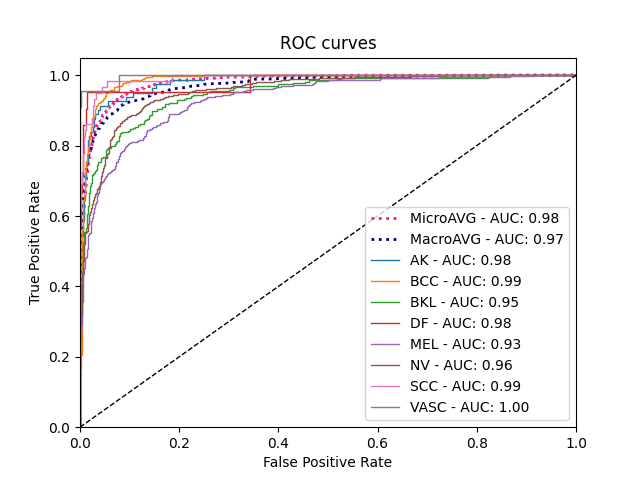}
  }
  \quad
  \subfigure[SSKD]{
  \includegraphics[width = 5cm]{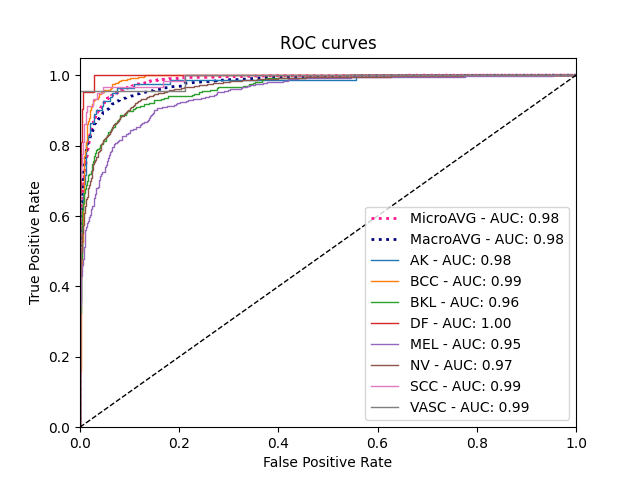}
  }
  \quad
  \subfigure[BLKD+DRKD]{
  \includegraphics[width = 5cm]{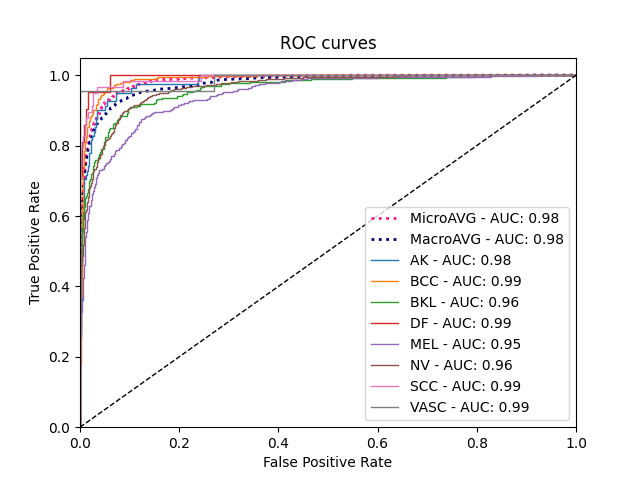}
  }
  \quad
  \subfigure[BLKD+CRKD]{
  \includegraphics[width = 5cm]{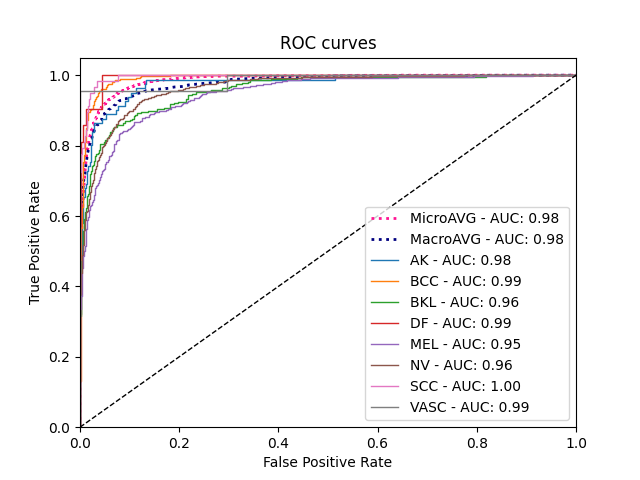}
  }
  \quad
  \subfigure[SSKD+DRKD]{
  \includegraphics[width = 5cm]{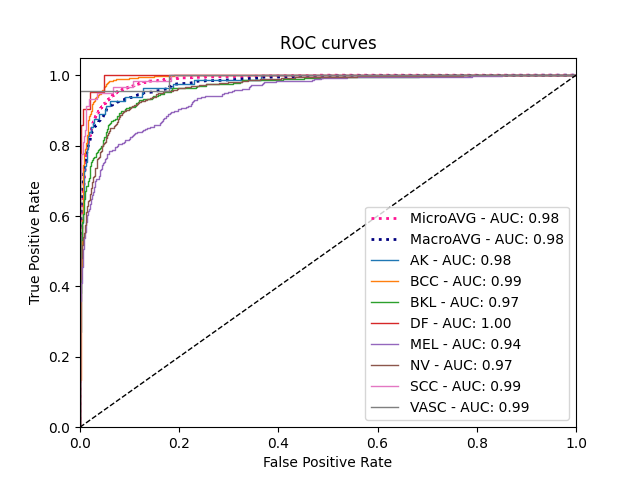}
  }
  \quad
  \subfigure[SSKD+CKD]{
  \includegraphics[width = 5cm]{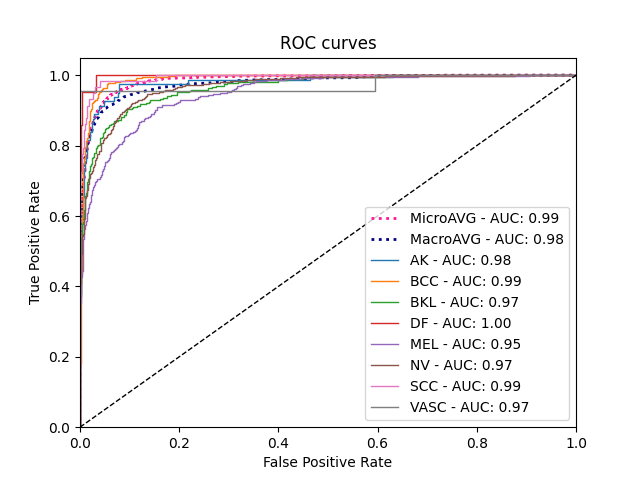}
  }
  \quad
  \subfigure[BLKD+DRKD+CRKD]{
  \includegraphics[width = 5cm]{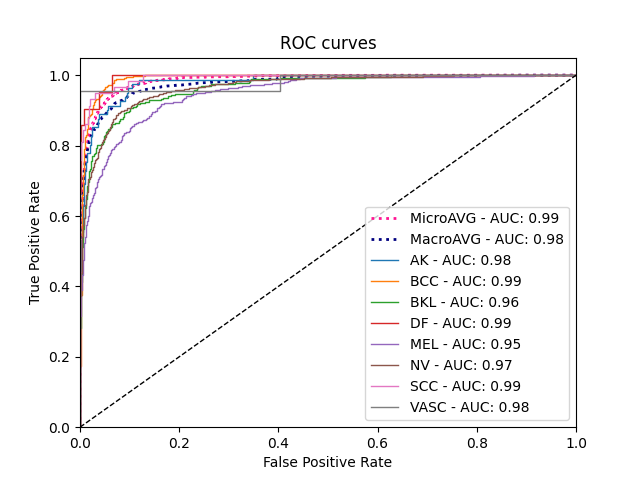}
  }
  \quad
  \subfigure[SSKD+DRKD+CRKD]{
  \includegraphics[width = 5cm]{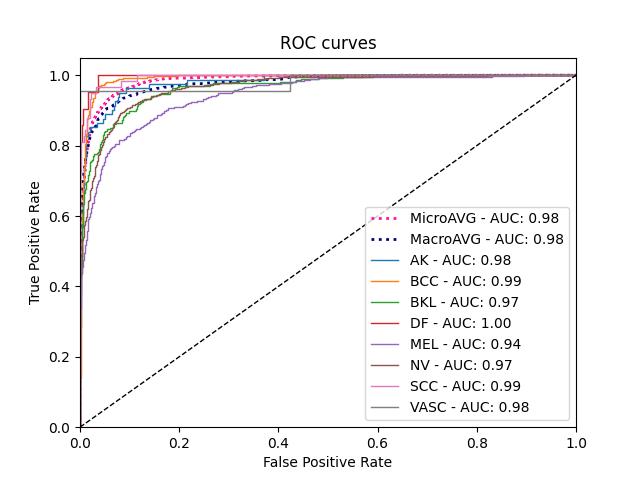}
  }
  \caption{Visualization of ROC curves from skin lesion classification models that are trained: without KD (as shown in in subfigures (a) -- (b)), and with different KDs (as shown in subfigures (c) -- (l)).}
  \label{fig:Roc}
\end{figure*}

To demonstrate the effectiveness of our proposed method, several popular representative knowledge distillation methods were listed for comparison. All these listed methods followed exactly the same training settings as introduced above to make sure the comparison was fair and the results were listed in Table~\ref{tbl2}, Figure~\ref{fig:conf_mat}, and Figure~\ref{fig:Roc}. As we can see, all knowledge distillation methods greatly improved the prediction performance of the student model, where the accuracy improvement ranged from 5\% to 9\% and the balanced accuracy improvement ranged from 4\% to 8\%. The increase in mean average precision even reached a maximum of 18\%. These results were comparable to the teacher model, and some parts were even higher than the teacher model. It proved that the KD method can improve the representative ability of the lightweight student model with the help of the teacher model in the classification task for multiple skin diseases via the dermoscopy imaging technique. Among all the methods listed in Table~\ref{tbl2}, we also found that our proposed method achieved the best performance in ACC, BACC, and mAP, and the second-best in AUC. Especially in the balanced accuracy, the obvious improvement indicated that our method is more suitable for the classification of multiple skin diseases with unbalanced sample numbers. In Figure~\ref{fig:conf_mat} and Figure~\ref{fig:Roc}, we presented the detailed confusion matrix of prediction and the ROC curves along with the AUC values for each diseases, which also supported our above analysis.

\subsection{Ablation studies}

As introduced in the methodology section, our proposed method was based on BLKD, while adding the distilled knowledge between instances and within the channels, along with a self-supervised learning strategy. Therefore, it was essential to explore the roles and effects of different components of the overall framework. In Table~\ref{tbl3}, we conducted ablation studies to compare the performance of different methods with different modules. In the training phase, we also fixed all the training optimization parameters and settings to make sure the comparison was fair. 

Firstly, we removed self-supervision mechanism and trained BLKD along with DRKD and CRKD separately as shown in the first two rows of the Table~\ref{tbl3}. Compared with BLKD, we found that both DRKD and CRKD methods consistently improved the accuracy, balanced accuracy, AUC, mAP values. Based on this result, we drew preliminary conclusions that the incorporation of relational-based KD can benefit the predictive ability of the student model by introducing more knowledge. Next, we involved the self-supervision in BLKD, which is noted as SSKD in the Table~\ref{tbl2} and Table~\ref{tbl3}. DRKD and CRKD methods were also considered individually in this scenario. According to the third and forth rows of the comparison, we noticed that the self-supervised method further improved the results while DRKD and CRKD methods were included, especially the ACC and BACC. This results demonstrated that while introducing DRKD and CRKD method to knowledge distillation, the adoption of the SSKD module can achieve further improvement of representative ability of the student model. Finally, we carried out another two sets of experiments that introduced both DRKD and CRKD modules and trained them with and without the self-supervision strategy to further validate our hypothesis. As shown in the last two rows of the Table~\ref{tbl3}, our jointly optimized knowledge distillation methods achieved the best performance with the highest prediction accuracy and balanced accuracy, which agreed with our preliminary conclusions mentioned above. Each component of our architecture contributed to the improved performance of the lightweight student model, thus making the proposed method more effective.

\subsection{Visualizations of class activation maps}

\begin{figure*}[htbp]
  \centering
  
  \subfigure[Original]{
  \includegraphics[width = 3cm,height = 2.2cm]{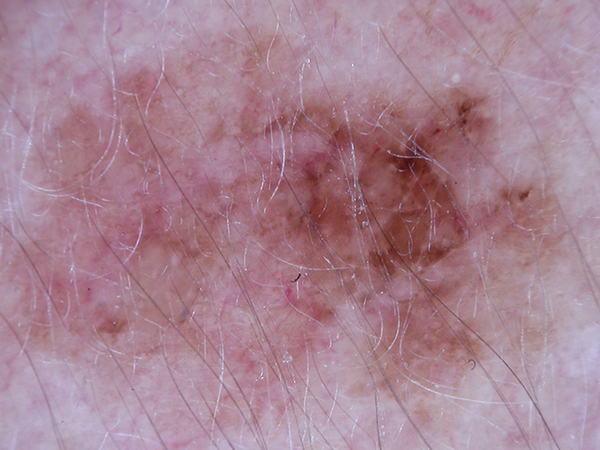}
  }
  \quad
  \subfigure[Teacher]{
  \includegraphics[width = 3cm,height = 2.2cm]{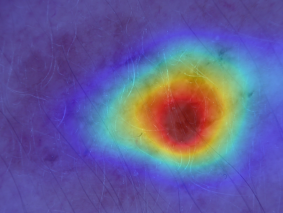}
  }
  \quad
  \subfigure[Student before SSD-KD]{
  \includegraphics[width = 3cm,height = 2.2cm]{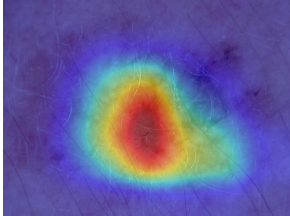}
  }
  \quad
  \subfigure[Student after SSD-KD]{
  \includegraphics[width = 3cm,height = 2.2cm]{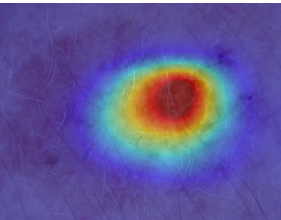}
  }
  
  \caption{Visualization of class activation mapping (CAM). The example images are correctly predicted by the teacher model and the student model only after implementing our proposed method. (a) is the original dermoscopy image while the other three subfigures (b)-(d) denote their corresponding CAM ouput from different models.}
  \label{fig:CAM}
\end{figure*}

In addition to the analysis and comparisons, we further observed the prediction performance of the test set samples before and after using our knowledge distillation method. We noticed that the skin lesion classes of many samples in the test set were correctly predicted by the teacher model and the student model with knowledge distillation, but are diagnosed as the wrong classes by the student model without the knowledge distillation. For example, 17 melanomas were predicted as benign keratoses by the student model without any knowledge distillation but they were classified correctly after the implementation of our KD method. To show how the KD methods helped the learning of the student model, we visualized the class activation mapping (CAM) from the teacher model, the student model before the implementation of KD, and the student model after using KD applying Grad-CAM \citep{selvaraju2017grad}, which was commonly used to locate discriminative regions for object detection and classification tasks. As presented in Figure~\ref{fig:CAM}, the located discriminative areas of student model after the KD implementation was more align with the teacher model's activation maps; Hence, it enhanced the performance of the student model.

\section{Discussion and future works}
The multi-class classification task has made great progress in recent years due to the in-depth research of medical imaging and deep learning technology. However, excellent model capabilities are often accompanied by complex structures and high training and storage costs, which limit their practical applications in the scenarios with restricted computational resources, such as portable devices. Correspondingly, although the small lightweight model can greatly reduce the computational cost, it cannot extract more representative features, so that the predictive ability of the model is degraded. For example, in this paper, both ResNet50 and MobileNetV2 are trained on the large-sacle dataset of ISIC. However, the prediction accuracy of ResNet50 can reach 82\%, while the lightweight MobileNet can only reach 75\%. It can be seen that using the KD methods to improve the learning performance of lightweight models while maintaining the complexity advantage is an important direction in both theoretical and application levels.

Considering the morphological characteristics of skin lesions, we added a number of modules based on the traditional KD framework, including the widely used DRKD and SSKD, as well as CRKD and weighted soften knowledge that were specifically designed for skin lesions classification via dermoscopy imaging. While the number of skin cancer lesions is far less than that of benign skin lesions, in our proposed method, we integrated weighted softened output and generic feature representations in a self-supervised manner, and also improved the ability of a lightweight model to obtain more useful information from classes with fewer subjects. Therefore, our proposed method can better help the lightweight student model to extract features, thereby achieving improved results. In addition, when we performed ablation studies, we also noticed that although the KD methods improved the student model substantially, the improvement brought by adding different modules to BLKD was not very obvious. To verify that the presented results are sufficiently convincing, we tried ten experiments under the same conditions and calculated the standard deviation of all calculated metrics. The standard deviation values were all less than 0.01, demonstrating the robustness of the trends shown by our results. In fact, further tuning of the hyper-parameters should make the model perform better when adding a combination of different modules, but in this paper we fixed the hyperparameter values of all modules for a completely fair comparison.

Additionally, after a series of pre-experiments, we selected the two most representative models, based on their distinct characteristics, as the teacher model and the student model. As the shown in table~\ref{tbl1}, our criteria for selecting models was to ensure that the teacher model has excellent performance, while the complexity of the student model is suitable for practical portable applications. Although we achieved our expectation of improving the predictive power of the student model, we strongly believed that this was only a starting point for this line of research. In our future works, we plan to further introduce other architectures, such as multi-teacher KD and KD with the help of the teaching assistant mechanism, so as to improve the generalizability of the obtained results. Moreover, our proposed method can also be applied to other medical problems with similar scenarios after fine-tuning, helping to improve lightweight models in other areas.

\section{Conclusion}

In this paper, we proposed an efficient knowledge distillation framework, called SSD-KD, which aimed to address the challenging task of multi-class classification of dermoscopic skin lesion imaging by lightweight deep learning models. Different from existing KD methods, SSD-KD considered both intra- and inter-instance based knowledge to guide the lightweight model capture richer structural knowledge that are important for skin lesion diagnosis. Additionally, the self-supervised auxiliary training strategy and weighted cross entropy loss were also involved to further boost its performance. 

We conducted experiments to evaluate our proposed method on the large-scale open-accessed dermoscopic skin lesion imaging dataset ISIC 2019, and we presented detailed comparisons with the existing KD methods. We also applied a CAM method to visualize the changes of activation maps generated by the models. Experiments demonstrated the effectiveness of our proposed SSD-KD with a promising result in multi-diseases classification. With low model complexity and storage requirements while improving predictive performance, our method could greatly assist in the promotion of deep learning techniques to clinical practices.

\printcredits

\bibliographystyle{cas-model2-names}

\bibliography{cas-refs}

\begin{thebibliography}{57}
\expandafter\ifx\csname natexlab\endcsname\relax\def\natexlab#1{#1}\fi
\providecommand{\url}[1]{\texttt{#1}}
\providecommand{\href}[2]{#2}
\providecommand{\path}[1]{#1}
\providecommand{\DOIprefix}{doi:}
\providecommand{\ArXivprefix}{arXiv:}
\providecommand{\URLprefix}{URL: }
\providecommand{\Pubmedprefix}{pmid:}
\providecommand{\doi}[1]{\href{http://dx.doi.org/#1}{\path{#1}}}
\providecommand{\Pubmed}[1]{\href{pmid:#1}{\path{#1}}}
\providecommand{\bibinfo}[2]{#2}
\ifx\xfnm\relax \def\xfnm[#1]{\unskip,\space#1}\fi
\bibitem[{Abbas and Celebi(2019)}]{abbas2019dermodeep}
\bibinfo{author}{Abbas, Q.}, \bibinfo{author}{Celebi, M.E.},
  \bibinfo{year}{2019}.
\newblock \bibinfo{title}{Dermodeep-a classification of melanoma-nevus skin
  lesions using multi-feature fusion of visual features and deep neural
  network}.
\newblock \bibinfo{journal}{Multimedia Tools and Applications}
  \bibinfo{volume}{78}, \bibinfo{pages}{23559--23580}.
\bibitem[{Abbasi et~al.(2021)Abbasi, Hajabdollahi, Khadivi, Karimi, Roshandel,
  Shirani and Samavi}]{abbasi2021classification}
\bibinfo{author}{Abbasi, S.}, \bibinfo{author}{Hajabdollahi, M.},
  \bibinfo{author}{Khadivi, P.}, \bibinfo{author}{Karimi, N.},
  \bibinfo{author}{Roshandel, R.}, \bibinfo{author}{Shirani, S.},
  \bibinfo{author}{Samavi, S.}, \bibinfo{year}{2021}.
\newblock \bibinfo{title}{Classification of diabetic retinopathy using
  unlabeled data and knowledge distillation}.
\newblock \bibinfo{journal}{Artificial Intelligence in Medicine}
  \bibinfo{volume}{121}, \bibinfo{pages}{102176}.
\bibitem[{Apalla et~al.(2017)Apalla, Lallas, Sotiriou, Lazaridou and
  Ioannides}]{apalla2017epidemiological}
\bibinfo{author}{Apalla, Z.}, \bibinfo{author}{Lallas, A.},
  \bibinfo{author}{Sotiriou, E.}, \bibinfo{author}{Lazaridou, E.},
  \bibinfo{author}{Ioannides, D.}, \bibinfo{year}{2017}.
\newblock \bibinfo{title}{Epidemiological trends in skin cancer}.
\newblock \bibinfo{journal}{Dermatology practical \& conceptual}
  \bibinfo{volume}{7}, \bibinfo{pages}{1}.
\bibitem[{Argenziano et~al.(1998)Argenziano, Fabbrocini, Carli, De~Giorgi,
  Sammarco and Delfino}]{argenziano1998epiluminescence}
\bibinfo{author}{Argenziano, G.}, \bibinfo{author}{Fabbrocini, G.},
  \bibinfo{author}{Carli, P.}, \bibinfo{author}{De~Giorgi, V.},
  \bibinfo{author}{Sammarco, E.}, \bibinfo{author}{Delfino, M.},
  \bibinfo{year}{1998}.
\newblock \bibinfo{title}{Epiluminescence microscopy for the diagnosis of
  doubtful melanocytic skin lesions: comparison of the abcd rule of
  dermatoscopy and a new 7-point checklist based on pattern analysis}.
\newblock \bibinfo{journal}{Archives of dermatology} \bibinfo{volume}{134},
  \bibinfo{pages}{1563--1570}.
\bibitem[{Back et~al.(2021)Back, Lee, Shin, Yu, Yuk, Jong, Ryu and
  Lee}]{back2021robust}
\bibinfo{author}{Back, S.}, \bibinfo{author}{Lee, S.}, \bibinfo{author}{Shin,
  S.}, \bibinfo{author}{Yu, Y.}, \bibinfo{author}{Yuk, T.},
  \bibinfo{author}{Jong, S.}, \bibinfo{author}{Ryu, S.}, \bibinfo{author}{Lee,
  K.}, \bibinfo{year}{2021}.
\newblock \bibinfo{title}{Robust skin disease classification by distilling deep
  neural network ensemble for the mobile diagnosis of herpes zoster}.
\newblock \bibinfo{journal}{IEEE Access} \bibinfo{volume}{9},
  \bibinfo{pages}{20156--20169}.
\bibitem[{Barata et~al.(2018)Barata, Celebi and Marques}]{barata2018survey}
\bibinfo{author}{Barata, C.}, \bibinfo{author}{Celebi, M.E.},
  \bibinfo{author}{Marques, J.S.}, \bibinfo{year}{2018}.
\newblock \bibinfo{title}{A survey of feature extraction in dermoscopy image
  analysis of skin cancer}.
\newblock \bibinfo{journal}{IEEE journal of biomedical and health informatics}
  \bibinfo{volume}{23}, \bibinfo{pages}{1096--1109}.
\bibitem[{Barata et~al.(2021)Barata, Celebi and
  Marques}]{barata2021explainable}
\bibinfo{author}{Barata, C.}, \bibinfo{author}{Celebi, M.E.},
  \bibinfo{author}{Marques, J.S.}, \bibinfo{year}{2021}.
\newblock \bibinfo{title}{Explainable skin lesion diagnosis using taxonomies}.
\newblock \bibinfo{journal}{Pattern Recognition} \bibinfo{volume}{110},
  \bibinfo{pages}{107413}.
\bibitem[{Bi et~al.(2020)Bi, Feng, Fulham and Kim}]{bi2020multi}
\bibinfo{author}{Bi, L.}, \bibinfo{author}{Feng, D.D.},
  \bibinfo{author}{Fulham, M.}, \bibinfo{author}{Kim, J.},
  \bibinfo{year}{2020}.
\newblock \bibinfo{title}{Multi-label classification of multi-modality skin
  lesion via hyper-connected convolutional neural network}.
\newblock \bibinfo{journal}{Pattern Recognition} \bibinfo{volume}{107},
  \bibinfo{pages}{107502}.
\bibitem[{Brinker et~al.(2018)Brinker, Hekler, Utikal, Grabe, Schadendorf,
  Klode, Berking, Steeb, Enk and Von~Kalle}]{brinker2018skin}
\bibinfo{author}{Brinker, T.J.}, \bibinfo{author}{Hekler, A.},
  \bibinfo{author}{Utikal, J.S.}, \bibinfo{author}{Grabe, N.},
  \bibinfo{author}{Schadendorf, D.}, \bibinfo{author}{Klode, J.},
  \bibinfo{author}{Berking, C.}, \bibinfo{author}{Steeb, T.},
  \bibinfo{author}{Enk, A.H.}, \bibinfo{author}{Von~Kalle, C.},
  \bibinfo{year}{2018}.
\newblock \bibinfo{title}{Skin cancer classification using convolutional neural
  networks: systematic review}.
\newblock \bibinfo{journal}{Journal of medical Internet research}
  \bibinfo{volume}{20}, \bibinfo{pages}{e11936}.
\bibitem[{Celebi et~al.(2019)Celebi, Codella and
  Halpern}]{celebi2019dermoscopy}
\bibinfo{author}{Celebi, M.E.}, \bibinfo{author}{Codella, N.},
  \bibinfo{author}{Halpern, A.}, \bibinfo{year}{2019}.
\newblock \bibinfo{title}{Dermoscopy image analysis: overview and future
  directions}.
\newblock \bibinfo{journal}{IEEE journal of biomedical and health informatics}
  \bibinfo{volume}{23}, \bibinfo{pages}{474--478}.
\bibitem[{Chen et~al.(2020)Chen, Kornblith, Norouzi and
  Hinton}]{chen2020simple}
\bibinfo{author}{Chen, T.}, \bibinfo{author}{Kornblith, S.},
  \bibinfo{author}{Norouzi, M.}, \bibinfo{author}{Hinton, G.},
  \bibinfo{year}{2020}.
\newblock \bibinfo{title}{A simple framework for contrastive learning of visual
  representations}, in: \bibinfo{booktitle}{International conference on machine
  learning}, \bibinfo{organization}{PMLR}. pp. \bibinfo{pages}{1597--1607}.
\bibitem[{Chen et~al.(2022)Chen, Gao, Li and Shen}]{chen2022lightweight}
\bibinfo{author}{Chen, W.}, \bibinfo{author}{Gao, L.}, \bibinfo{author}{Li,
  X.}, \bibinfo{author}{Shen, W.}, \bibinfo{year}{2022}.
\newblock \bibinfo{title}{Lightweight convolutional neural network with
  knowledge distillation for cervical cells classification}.
\newblock \bibinfo{journal}{Biomedical Signal Processing and Control}
  \bibinfo{volume}{71}, \bibinfo{pages}{103177}.
\bibitem[{Codella et~al.(2018)Codella, Gutman, Celebi, Helba, Marchetti, Dusza,
  Kalloo, Liopyris, Mishra, Kittler et~al.}]{codella2018skin}
\bibinfo{author}{Codella, N.C.}, \bibinfo{author}{Gutman, D.},
  \bibinfo{author}{Celebi, M.E.}, \bibinfo{author}{Helba, B.},
  \bibinfo{author}{Marchetti, M.A.}, \bibinfo{author}{Dusza, S.W.},
  \bibinfo{author}{Kalloo, A.}, \bibinfo{author}{Liopyris, K.},
  \bibinfo{author}{Mishra, N.}, \bibinfo{author}{Kittler, H.}, et~al.,
  \bibinfo{year}{2018}.
\newblock \bibinfo{title}{Skin lesion analysis toward melanoma detection: A
  challenge at the 2017 international symposium on biomedical imaging (isbi),
  hosted by the international skin imaging collaboration (isic)}, in:
  \bibinfo{booktitle}{2018 IEEE 15th international symposium on biomedical
  imaging (ISBI 2018)}, \bibinfo{organization}{IEEE}. pp.
  \bibinfo{pages}{168--172}.
\bibitem[{Combalia et~al.(2019)Combalia, Codella, Rotemberg, Helba, Vilaplana,
  Reiter, Carrera, Barreiro, Halpern, Puig et~al.}]{combalia2019bcn20000}
\bibinfo{author}{Combalia, M.}, \bibinfo{author}{Codella, N.C.},
  \bibinfo{author}{Rotemberg, V.}, \bibinfo{author}{Helba, B.},
  \bibinfo{author}{Vilaplana, V.}, \bibinfo{author}{Reiter, O.},
  \bibinfo{author}{Carrera, C.}, \bibinfo{author}{Barreiro, A.},
  \bibinfo{author}{Halpern, A.C.}, \bibinfo{author}{Puig, S.}, et~al.,
  \bibinfo{year}{2019}.
\newblock \bibinfo{title}{Bcn20000: Dermoscopic lesions in the wild}.
\newblock \bibinfo{journal}{arXiv preprint arXiv:1908.02288} .
\bibitem[{Deng et~al.(2009)Deng, Dong, Socher, Li, Li and
  Fei-Fei}]{deng2009imagenet}
\bibinfo{author}{Deng, J.}, \bibinfo{author}{Dong, W.},
  \bibinfo{author}{Socher, R.}, \bibinfo{author}{Li, L.J.},
  \bibinfo{author}{Li, K.}, \bibinfo{author}{Fei-Fei, L.},
  \bibinfo{year}{2009}.
\newblock \bibinfo{title}{Imagenet: A large-scale hierarchical image database},
  in: \bibinfo{booktitle}{2009 IEEE conference on computer vision and pattern
  recognition}, \bibinfo{organization}{Ieee}. pp. \bibinfo{pages}{248--255}.
\bibitem[{Ding et~al.(2021a)Ding, Wang, Yuan, Jiang, Wang, Huang and
  Wang}]{ding2021towards}
\bibinfo{author}{Ding, L.}, \bibinfo{author}{Wang, Y.}, \bibinfo{author}{Yuan,
  K.}, \bibinfo{author}{Jiang, M.}, \bibinfo{author}{Wang, P.},
  \bibinfo{author}{Huang, H.}, \bibinfo{author}{Wang, Z.J.},
  \bibinfo{year}{2021}a.
\newblock \bibinfo{title}{Towards universal physical attacks on single object
  tracking}, in: \bibinfo{booktitle}{Proceedings of the AAAI Conference on
  Artificial Intelligence}, pp. \bibinfo{pages}{1236--1245}.
\bibitem[{Ding et~al.(2021b)Ding, Wang, Wang and Welch}]{ding2021efficient}
\bibinfo{author}{Ding, X.}, \bibinfo{author}{Wang, Y.}, \bibinfo{author}{Wang,
  Z.J.}, \bibinfo{author}{Welch, W.J.}, \bibinfo{year}{2021}b.
\newblock \bibinfo{title}{Efficient subsampling for generating high-quality
  images from conditional generative adversarial networks}.
\newblock \bibinfo{journal}{arXiv preprint arXiv:2103.11166} .
\bibitem[{Ding et~al.(2021c)Ding, Wang, Xu, Wang and
  Welch}]{ding2021distilling}
\bibinfo{author}{Ding, X.}, \bibinfo{author}{Wang, Y.}, \bibinfo{author}{Xu,
  Z.}, \bibinfo{author}{Wang, Z.J.}, \bibinfo{author}{Welch, W.J.},
  \bibinfo{year}{2021}c.
\newblock \bibinfo{title}{Distilling and transferring knowledge via
  cgan-generated samples for image classification and regression}.
\newblock \bibinfo{journal}{arXiv preprint arXiv:2104.03164} .
\bibitem[{Ding et~al.(2021d)Ding, Wang, Xu, Welch and Wang}]{ding2020ccgan}
\bibinfo{author}{Ding, X.}, \bibinfo{author}{Wang, Y.}, \bibinfo{author}{Xu,
  Z.}, \bibinfo{author}{Welch, W.J.}, \bibinfo{author}{Wang, Z.J.},
  \bibinfo{year}{2021}d.
\newblock \bibinfo{title}{Ccgan: Continuous conditional generative adversarial
  networks for image generation}, in: \bibinfo{booktitle}{International
  Conference on Learning Representations}, pp. \bibinfo{pages}{1--10}.
\bibitem[{Esteva et~al.(2017)Esteva, Kuprel, Novoa, Ko, Swetter, Blau and
  Thrun}]{esteva2017dermatologist}
\bibinfo{author}{Esteva, A.}, \bibinfo{author}{Kuprel, B.},
  \bibinfo{author}{Novoa, R.A.}, \bibinfo{author}{Ko, J.},
  \bibinfo{author}{Swetter, S.M.}, \bibinfo{author}{Blau, H.M.},
  \bibinfo{author}{Thrun, S.}, \bibinfo{year}{2017}.
\newblock \bibinfo{title}{Dermatologist-level classification of skin cancer
  with deep neural networks}.
\newblock \bibinfo{journal}{nature} \bibinfo{volume}{542},
  \bibinfo{pages}{115--118}.
\bibitem[{Geirhos et~al.(2018)Geirhos, Rubisch, Michaelis, Bethge, Wichmann and
  Brendel}]{geirhos2018imagenet}
\bibinfo{author}{Geirhos, R.}, \bibinfo{author}{Rubisch, P.},
  \bibinfo{author}{Michaelis, C.}, \bibinfo{author}{Bethge, M.},
  \bibinfo{author}{Wichmann, F.A.}, \bibinfo{author}{Brendel, W.},
  \bibinfo{year}{2018}.
\newblock \bibinfo{title}{Imagenet-trained cnns are biased towards texture;
  increasing shape bias improves accuracy and robustness}.
\newblock \bibinfo{journal}{arXiv preprint arXiv:1811.12231} .
\bibitem[{Gessert et~al.(2020)Gessert, Nielsen, Shaikh, Werner and
  Schlaefer}]{gessert2020skin}
\bibinfo{author}{Gessert, N.}, \bibinfo{author}{Nielsen, M.},
  \bibinfo{author}{Shaikh, M.}, \bibinfo{author}{Werner, R.},
  \bibinfo{author}{Schlaefer, A.}, \bibinfo{year}{2020}.
\newblock \bibinfo{title}{Skin lesion classification using ensembles of
  multi-resolution efficientnets with meta data}.
\newblock \bibinfo{journal}{MethodsX} \bibinfo{volume}{7},
  \bibinfo{pages}{100864}.
\bibitem[{Gou et~al.(2021)Gou, Yu, Maybank and Tao}]{gou2021knowledge}
\bibinfo{author}{Gou, J.}, \bibinfo{author}{Yu, B.}, \bibinfo{author}{Maybank,
  S.J.}, \bibinfo{author}{Tao, D.}, \bibinfo{year}{2021}.
\newblock \bibinfo{title}{Knowledge distillation: A survey}.
\newblock \bibinfo{journal}{International Journal of Computer Vision}
  \bibinfo{volume}{129}, \bibinfo{pages}{1789--1819}.
\bibitem[{He et~al.(2020)He, Fan, Wu, Xie and Girshick}]{he2020momentum}
\bibinfo{author}{He, K.}, \bibinfo{author}{Fan, H.}, \bibinfo{author}{Wu, Y.},
  \bibinfo{author}{Xie, S.}, \bibinfo{author}{Girshick, R.},
  \bibinfo{year}{2020}.
\newblock \bibinfo{title}{Momentum contrast for unsupervised visual
  representation learning}, in: \bibinfo{booktitle}{Proceedings of the IEEE/CVF
  conference on computer vision and pattern recognition}, pp.
  \bibinfo{pages}{9729--9738}.
\bibitem[{He et~al.(2016)He, Zhang, Ren and Sun}]{he2016deep}
\bibinfo{author}{He, K.}, \bibinfo{author}{Zhang, X.}, \bibinfo{author}{Ren,
  S.}, \bibinfo{author}{Sun, J.}, \bibinfo{year}{2016}.
\newblock \bibinfo{title}{Deep residual learning for image recognition}, in:
  \bibinfo{booktitle}{Proceedings of the IEEE conference on computer vision and
  pattern recognition}, pp. \bibinfo{pages}{770--778}.
\bibitem[{Hinton et~al.(2015)Hinton, Vinyals, Dean
  et~al.}]{hinton2015distilling}
\bibinfo{author}{Hinton, G.}, \bibinfo{author}{Vinyals, O.},
  \bibinfo{author}{Dean, J.}, et~al., \bibinfo{year}{2015}.
\newblock \bibinfo{title}{Distilling the knowledge in a neural network}.
\newblock \bibinfo{journal}{arXiv preprint arXiv:1503.02531}
  \bibinfo{volume}{2}.
\bibitem[{Huber(1992)}]{huber1992robust}
\bibinfo{author}{Huber, P.J.}, \bibinfo{year}{1992}.
\newblock \bibinfo{title}{Robust estimation of a location parameter}, in:
  \bibinfo{booktitle}{Breakthroughs in statistics}.
  \bibinfo{publisher}{Springer}, pp. \bibinfo{pages}{492--518}.
\bibitem[{Jing and Tian(2020)}]{jing2020self}
\bibinfo{author}{Jing, L.}, \bibinfo{author}{Tian, Y.}, \bibinfo{year}{2020}.
\newblock \bibinfo{title}{Self-supervised visual feature learning with deep
  neural networks: A survey}.
\newblock \bibinfo{journal}{IEEE transactions on pattern analysis and machine
  intelligence} \bibinfo{volume}{43}, \bibinfo{pages}{4037--4058}.
\bibitem[{Johnson et~al.(2016)Johnson, Alahi and
  Fei-Fei}]{johnson2016perceptual}
\bibinfo{author}{Johnson, J.}, \bibinfo{author}{Alahi, A.},
  \bibinfo{author}{Fei-Fei, L.}, \bibinfo{year}{2016}.
\newblock \bibinfo{title}{Perceptual losses for real-time style transfer and
  super-resolution}, in: \bibinfo{booktitle}{European conference on computer
  vision}, \bibinfo{organization}{Springer}. pp. \bibinfo{pages}{694--711}.
\bibitem[{Kawahara et~al.(2016)Kawahara, BenTaieb and
  Hamarneh}]{kawahara2016deep}
\bibinfo{author}{Kawahara, J.}, \bibinfo{author}{BenTaieb, A.},
  \bibinfo{author}{Hamarneh, G.}, \bibinfo{year}{2016}.
\newblock \bibinfo{title}{Deep features to classify skin lesions}, in:
  \bibinfo{booktitle}{2016 IEEE 13th international symposium on biomedical
  imaging (ISBI)}, \bibinfo{organization}{IEEE}. pp.
  \bibinfo{pages}{1397--1400}.
\bibitem[{Kawahara et~al.(2018)Kawahara, Daneshvar, Argenziano and
  Hamarneh}]{kawahara2018seven}
\bibinfo{author}{Kawahara, J.}, \bibinfo{author}{Daneshvar, S.},
  \bibinfo{author}{Argenziano, G.}, \bibinfo{author}{Hamarneh, G.},
  \bibinfo{year}{2018}.
\newblock \bibinfo{title}{Seven-point checklist and skin lesion classification
  using multitask multimodal neural nets}.
\newblock \bibinfo{journal}{IEEE journal of biomedical and health informatics}
  \bibinfo{volume}{23}, \bibinfo{pages}{538--546}.
\bibitem[{Lin et~al.(2014)Lin, Chen and Yan}]{lin2014network}
\bibinfo{author}{Lin, M.}, \bibinfo{author}{Chen, Q.}, \bibinfo{author}{Yan,
  S.}, \bibinfo{year}{2014}.
\newblock \bibinfo{title}{Network in network}, in:
  \bibinfo{booktitle}{International Conference on Learning Representations},
  pp. \bibinfo{pages}{1--10}.
\bibitem[{Litjens et~al.(2017)Litjens, Kooi, Bejnordi, Setio, Ciompi,
  Ghafoorian, Van Der~Laak, Van~Ginneken and S{\'a}nchez}]{litjens2017survey}
\bibinfo{author}{Litjens, G.}, \bibinfo{author}{Kooi, T.},
  \bibinfo{author}{Bejnordi, B.E.}, \bibinfo{author}{Setio, A.A.A.},
  \bibinfo{author}{Ciompi, F.}, \bibinfo{author}{Ghafoorian, M.},
  \bibinfo{author}{Van Der~Laak, J.A.}, \bibinfo{author}{Van~Ginneken, B.},
  \bibinfo{author}{S{\'a}nchez, C.I.}, \bibinfo{year}{2017}.
\newblock \bibinfo{title}{A survey on deep learning in medical image analysis}.
\newblock \bibinfo{journal}{Medical image analysis} \bibinfo{volume}{42},
  \bibinfo{pages}{60--88}.
\bibitem[{Liu et~al.(2022)Liu, HaoChen, Gaidon and Ma}]{liu2022self}
\bibinfo{author}{Liu, H.}, \bibinfo{author}{HaoChen, J.Z.},
  \bibinfo{author}{Gaidon, A.}, \bibinfo{author}{Ma, T.}, \bibinfo{year}{2022}.
\newblock \bibinfo{title}{Self-supervised learning is more robust to dataset
  imbalance}, in: \bibinfo{booktitle}{International Conference on Learning
  Representations}, pp. \bibinfo{pages}{1--8}.
\bibitem[{Liu et~al.(2019)Liu, Cao, Li, Yuan, Hu, Li and
  Duan}]{liu2019knowledge}
\bibinfo{author}{Liu, Y.}, \bibinfo{author}{Cao, J.}, \bibinfo{author}{Li, B.},
  \bibinfo{author}{Yuan, C.}, \bibinfo{author}{Hu, W.}, \bibinfo{author}{Li,
  Y.}, \bibinfo{author}{Duan, Y.}, \bibinfo{year}{2019}.
\newblock \bibinfo{title}{Knowledge distillation via instance relationship
  graph}, in: \bibinfo{booktitle}{Proceedings of the IEEE/CVF Conference on
  Computer Vision and Pattern Recognition}, pp. \bibinfo{pages}{7096--7104}.
\bibitem[{Nachbar et~al.(1994)Nachbar, Stolz, Merkle, Cognetta, Vogt,
  Landthaler, Bilek, Braun-Falco and Plewig}]{nachbar1994abcd}
\bibinfo{author}{Nachbar, F.}, \bibinfo{author}{Stolz, W.},
  \bibinfo{author}{Merkle, T.}, \bibinfo{author}{Cognetta, A.B.},
  \bibinfo{author}{Vogt, T.}, \bibinfo{author}{Landthaler, M.},
  \bibinfo{author}{Bilek, P.}, \bibinfo{author}{Braun-Falco, O.},
  \bibinfo{author}{Plewig, G.}, \bibinfo{year}{1994}.
\newblock \bibinfo{title}{The abcd rule of dermatoscopy: high prospective value
  in the diagnosis of doubtful melanocytic skin lesions}.
\newblock \bibinfo{journal}{Journal of the American Academy of Dermatology}
  \bibinfo{volume}{30}, \bibinfo{pages}{551--559}.
\bibitem[{Van~den Oord et~al.(2018)Van~den Oord, Li and
  Vinyals}]{van2018representation}
\bibinfo{author}{Van~den Oord, A.}, \bibinfo{author}{Li, Y.},
  \bibinfo{author}{Vinyals, O.}, \bibinfo{year}{2018}.
\newblock \bibinfo{title}{Representation learning with contrastive predictive
  coding}.
\newblock \bibinfo{journal}{arXiv e-prints} , \bibinfo{pages}{arXiv--1807}.
\bibitem[{Pacheco and Krohling(2021)}]{pacheco2021attention}
\bibinfo{author}{Pacheco, A.G.}, \bibinfo{author}{Krohling, R.A.},
  \bibinfo{year}{2021}.
\newblock \bibinfo{title}{An attention-based mechanism to combine images and
  metadata in deep learning models applied to skin cancer classification}.
\newblock \bibinfo{journal}{IEEE journal of biomedical and health informatics}
  \bibinfo{volume}{25}, \bibinfo{pages}{3554--3563}.
\bibitem[{Park et~al.(2019)Park, Kim, Lu and Cho}]{park2019relational}
\bibinfo{author}{Park, W.}, \bibinfo{author}{Kim, D.}, \bibinfo{author}{Lu,
  Y.}, \bibinfo{author}{Cho, M.}, \bibinfo{year}{2019}.
\newblock \bibinfo{title}{Relational knowledge distillation}, in:
  \bibinfo{booktitle}{Proceedings of the IEEE/CVF Conference on Computer Vision
  and Pattern Recognition}, pp. \bibinfo{pages}{3967--3976}.
\bibitem[{Paszke et~al.(2019)Paszke, Gross, Massa, Lerer, Bradbury, Chanan,
  Killeen, Lin, Gimelshein, Antiga et~al.}]{paszke2019pytorch}
\bibinfo{author}{Paszke, A.}, \bibinfo{author}{Gross, S.},
  \bibinfo{author}{Massa, F.}, \bibinfo{author}{Lerer, A.},
  \bibinfo{author}{Bradbury, J.}, \bibinfo{author}{Chanan, G.},
  \bibinfo{author}{Killeen, T.}, \bibinfo{author}{Lin, Z.},
  \bibinfo{author}{Gimelshein, N.}, \bibinfo{author}{Antiga, L.}, et~al.,
  \bibinfo{year}{2019}.
\newblock \bibinfo{title}{Pytorch: An imperative style, high-performance deep
  learning library}.
\newblock \bibinfo{journal}{Advances in neural information processing systems}
  \bibinfo{volume}{32}.
\bibitem[{Peng et~al.(2019)Peng, Jin, Liu, Li, Wu, Liu, Zhou and
  Zhang}]{peng2019correlation}
\bibinfo{author}{Peng, B.}, \bibinfo{author}{Jin, X.}, \bibinfo{author}{Liu,
  J.}, \bibinfo{author}{Li, D.}, \bibinfo{author}{Wu, Y.},
  \bibinfo{author}{Liu, Y.}, \bibinfo{author}{Zhou, S.},
  \bibinfo{author}{Zhang, Z.}, \bibinfo{year}{2019}.
\newblock \bibinfo{title}{Correlation congruence for knowledge distillation},
  in: \bibinfo{booktitle}{Proceedings of the IEEE/CVF International Conference
  on Computer Vision}, pp. \bibinfo{pages}{5007--5016}.
\bibitem[{Qin et~al.(2021)Qin, Bu, Liu, Shen, Zhou, Gu, Wang, Wu and
  Dai}]{qin2021efficient}
\bibinfo{author}{Qin, D.}, \bibinfo{author}{Bu, J.J.}, \bibinfo{author}{Liu,
  Z.}, \bibinfo{author}{Shen, X.}, \bibinfo{author}{Zhou, S.},
  \bibinfo{author}{Gu, J.J.}, \bibinfo{author}{Wang, Z.H.},
  \bibinfo{author}{Wu, L.}, \bibinfo{author}{Dai, H.F.}, \bibinfo{year}{2021}.
\newblock \bibinfo{title}{Efficient medical image segmentation based on
  knowledge distillation}.
\newblock \bibinfo{journal}{IEEE Transactions on Medical Imaging}
  \bibinfo{volume}{40}, \bibinfo{pages}{3820--3831}.
\bibitem[{Raghu et~al.(2019)Raghu, Zhang, Kleinberg and
  Bengio}]{raghu2019transfusion}
\bibinfo{author}{Raghu, M.}, \bibinfo{author}{Zhang, C.},
  \bibinfo{author}{Kleinberg, J.}, \bibinfo{author}{Bengio, S.},
  \bibinfo{year}{2019}.
\newblock \bibinfo{title}{Transfusion: Understanding transfer learning for
  medical imaging}.
\newblock \bibinfo{journal}{Advances in neural information processing systems}
  \bibinfo{volume}{32}.
\bibitem[{Romero et~al.(2015)Romero, Ballas, Kahou, Chassang, Gatta and
  Bengio}]{romero2015fitnets}
\bibinfo{author}{Romero, A.}, \bibinfo{author}{Ballas, N.},
  \bibinfo{author}{Kahou, S.E.}, \bibinfo{author}{Chassang, A.},
  \bibinfo{author}{Gatta, C.}, \bibinfo{author}{Bengio, Y.},
  \bibinfo{year}{2015}.
\newblock \bibinfo{title}{Fitnets: Hints for thin deep nets}.
\newblock \bibinfo{journal}{International Conference on Learning
  Representations} .
\bibitem[{Sandler et~al.(2018)Sandler, Howard, Zhu, Zhmoginov and
  Chen}]{sandler2018mobilenetv2}
\bibinfo{author}{Sandler, M.}, \bibinfo{author}{Howard, A.},
  \bibinfo{author}{Zhu, M.}, \bibinfo{author}{Zhmoginov, A.},
  \bibinfo{author}{Chen, L.C.}, \bibinfo{year}{2018}.
\newblock \bibinfo{title}{Mobilenetv2: Inverted residuals and linear
  bottlenecks}, in: \bibinfo{booktitle}{Proceedings of the IEEE conference on
  computer vision and pattern recognition}, pp. \bibinfo{pages}{4510--4520}.
\bibitem[{Selvaraju et~al.(2017)Selvaraju, Cogswell, Das, Vedantam, Parikh and
  Batra}]{selvaraju2017grad}
\bibinfo{author}{Selvaraju, R.R.}, \bibinfo{author}{Cogswell, M.},
  \bibinfo{author}{Das, A.}, \bibinfo{author}{Vedantam, R.},
  \bibinfo{author}{Parikh, D.}, \bibinfo{author}{Batra, D.},
  \bibinfo{year}{2017}.
\newblock \bibinfo{title}{Grad-cam: Visual explanations from deep networks via
  gradient-based localization}, in: \bibinfo{booktitle}{Proceedings of the IEEE
  international conference on computer vision}, pp. \bibinfo{pages}{618--626}.
\bibitem[{Srinivasu et~al.(2021)Srinivasu, SivaSai, Ijaz, Bhoi, Kim and
  Kang}]{srinivasu2021classification}
\bibinfo{author}{Srinivasu, P.N.}, \bibinfo{author}{SivaSai, J.G.},
  \bibinfo{author}{Ijaz, M.F.}, \bibinfo{author}{Bhoi, A.K.},
  \bibinfo{author}{Kim, W.}, \bibinfo{author}{Kang, J.J.},
  \bibinfo{year}{2021}.
\newblock \bibinfo{title}{Classification of skin disease using deep learning
  neural networks with mobilenet v2 and lstm}.
\newblock \bibinfo{journal}{Sensors} \bibinfo{volume}{21},
  \bibinfo{pages}{2852}.
\bibitem[{Tang et~al.(2022)Tang, Yan, Nan, Xiang, Krammer and
  Lasser}]{tang2022fusionm4net}
\bibinfo{author}{Tang, P.}, \bibinfo{author}{Yan, X.}, \bibinfo{author}{Nan,
  Y.}, \bibinfo{author}{Xiang, S.}, \bibinfo{author}{Krammer, S.},
  \bibinfo{author}{Lasser, T.}, \bibinfo{year}{2022}.
\newblock \bibinfo{title}{Fusionm4net: A multi-stage multi-modal learning
  algorithm for multi-label skin lesion classification}.
\newblock \bibinfo{journal}{Medical Image Analysis} \bibinfo{volume}{76},
  \bibinfo{pages}{102307}.
\bibitem[{To{\u{g}}a{\c{c}}ar et~al.(2021)To{\u{g}}a{\c{c}}ar, C{\"o}mert and
  Ergen}]{tougaccar2021intelligent}
\bibinfo{author}{To{\u{g}}a{\c{c}}ar, M.}, \bibinfo{author}{C{\"o}mert, Z.},
  \bibinfo{author}{Ergen, B.}, \bibinfo{year}{2021}.
\newblock \bibinfo{title}{Intelligent skin cancer detection applying
  autoencoder, mobilenetv2 and spiking neural networks}.
\newblock \bibinfo{journal}{Chaos, Solitons \& Fractals} \bibinfo{volume}{144},
  \bibinfo{pages}{110714}.
\bibitem[{Tschandl et~al.(2018)Tschandl, Rosendahl and
  Kittler}]{tschandl2018ham10000}
\bibinfo{author}{Tschandl, P.}, \bibinfo{author}{Rosendahl, C.},
  \bibinfo{author}{Kittler, H.}, \bibinfo{year}{2018}.
\newblock \bibinfo{title}{The ham10000 dataset, a large collection of
  multi-source dermatoscopic images of common pigmented skin lesions}.
\newblock \bibinfo{journal}{Scientific data} \bibinfo{volume}{5},
  \bibinfo{pages}{1--9}.
\bibitem[{Tung and Mori(2019)}]{tung2019similarity}
\bibinfo{author}{Tung, F.}, \bibinfo{author}{Mori, G.}, \bibinfo{year}{2019}.
\newblock \bibinfo{title}{Similarity-preserving knowledge distillation}, in:
  \bibinfo{booktitle}{Proceedings of the IEEE/CVF International Conference on
  Computer Vision}, pp. \bibinfo{pages}{1365--1374}.
\bibitem[{Wang and Yoon(2021)}]{wang2021knowledge}
\bibinfo{author}{Wang, L.}, \bibinfo{author}{Yoon, K.J.}, \bibinfo{year}{2021}.
\newblock \bibinfo{title}{Knowledge distillation and student-teacher learning
  for visual intelligence: A review and new outlooks}.
\newblock \bibinfo{journal}{IEEE Transactions on Pattern Analysis and Machine
  Intelligence} .
\bibitem[{Wang et~al.(2021)Wang, Cai, Louie, Wang and
  Lee}]{wang2021incorporating}
\bibinfo{author}{Wang, Y.}, \bibinfo{author}{Cai, J.}, \bibinfo{author}{Louie,
  D.C.}, \bibinfo{author}{Wang, Z.J.}, \bibinfo{author}{Lee, T.K.},
  \bibinfo{year}{2021}.
\newblock \bibinfo{title}{Incorporating clinical knowledge with constrained
  classifier chain into a multimodal deep network for melanoma detection}.
\newblock \bibinfo{journal}{Computers in Biology and Medicine}
  \bibinfo{volume}{137}, \bibinfo{pages}{104812}.
\bibitem[{Xie et~al.(2021)Xie, Niu, Liu, Chen, Tang and Yu}]{xie2021survey}
\bibinfo{author}{Xie, X.}, \bibinfo{author}{Niu, J.}, \bibinfo{author}{Liu,
  X.}, \bibinfo{author}{Chen, Z.}, \bibinfo{author}{Tang, S.},
  \bibinfo{author}{Yu, S.}, \bibinfo{year}{2021}.
\newblock \bibinfo{title}{A survey on incorporating domain knowledge into deep
  learning for medical image analysis}.
\newblock \bibinfo{journal}{Medical Image Analysis} \bibinfo{volume}{69},
  \bibinfo{pages}{101985}.
\bibitem[{Xu et~al.(2020)Xu, Liu, Li and Loy}]{xu2020knowledge}
\bibinfo{author}{Xu, G.}, \bibinfo{author}{Liu, Z.}, \bibinfo{author}{Li, X.},
  \bibinfo{author}{Loy, C.C.}, \bibinfo{year}{2020}.
\newblock \bibinfo{title}{Knowledge distillation meets self-supervision}, in:
  \bibinfo{booktitle}{European Conference on Computer Vision},
  \bibinfo{organization}{Springer}. pp. \bibinfo{pages}{588--604}.
\bibitem[{Yim et~al.(2017)Yim, Joo, Bae and Kim}]{yim2017gift}
\bibinfo{author}{Yim, J.}, \bibinfo{author}{Joo, D.}, \bibinfo{author}{Bae,
  J.}, \bibinfo{author}{Kim, J.}, \bibinfo{year}{2017}.
\newblock \bibinfo{title}{A gift from knowledge distillation: Fast
  optimization, network minimization and transfer learning}, in:
  \bibinfo{booktitle}{Proceedings of the IEEE Conference on Computer Vision and
  Pattern Recognition}, pp. \bibinfo{pages}{4133--4141}.
\bibitem[{Zhang et~al.(2019)Zhang, Xie, Wu and Xia}]{zhang2019medical}
\bibinfo{author}{Zhang, J.}, \bibinfo{author}{Xie, Y.}, \bibinfo{author}{Wu,
  Q.}, \bibinfo{author}{Xia, Y.}, \bibinfo{year}{2019}.
\newblock \bibinfo{title}{Medical image classification using synergic deep
  learning}.
\newblock \bibinfo{journal}{Medical image analysis} \bibinfo{volume}{54},
  \bibinfo{pages}{10--19}.

\end{thebibliography}



\end{document}